\PassOptionsToPackage{pdfencoding=auto, psdextra}{hyperref}
\documentclass[final,1p,times]{elsarticle}
\usepackage{placeins}
\usepackage{amssymb}
\usepackage{amsmath}
\usepackage{amsthm}
\usepackage{xcolor}
\usepackage{graphicx}
\usepackage{hyperref}
\hypersetup{
    pdfencoding=auto,
    psdextra,
    colorlinks=true,
    linkcolor=blue,
    citecolor=blue,
    urlcolor=blue,
    pdfauthor={Ali Vefghi, Zahed Rahmati},
    pdftitle={Where Black-box Drug-Target Interaction Prediction Models Look: Cross-Method Explainability},
    pdfsubject={Subject},
    pdfkeywords={DTI, DTA, XAI}
}
\usepackage{multirow}

\usepackage{url}
\usepackage{booktabs}
\usepackage{xcolor}
\usepackage{array}
\usepackage{makecell}
\usepackage{tabularx}
\journal{Journal of ?}

\begin{document}
	
\begin{frontmatter}

\title{Where Black-box Drug-Target Interaction Prediction Models Look: Cross-Method Explainability}

        \author[label1]{Ali Vefghi}
        \ead{alivefghi@aut.ac.ir}
   \author[label1]{Zahed Rahmati\corref{cor1}} 
        \ead{zrahmati@aut.ac.ir}
    \author[label1]{Mohammad Akbari} 
        \ead{akbari.ma@aut.ac.ir}
    \affiliation[label1]{organization={Department of Mathematics and Computer Science,  Amirkabir University of Technology}, city={Tehran}, country={Iran}}
      
		\cortext[cor1]{Corresponding author}

\begin{abstract}
Drug-target interaction (DTI) and affinity (DTA) predictors increasingly achieve strong benchmark scores, yet their internal use of sequence, fingerprint, and graph features often remains opaque. 
We present an interpretability audit of BridgeDPI architecture on three different datasets including Gao, Human, and C.elegans. 
This study combines gradient-based attributions—integrated gradients, saliency, layer-wise relevance propagation, SmoothGrad, and SmoothGrad-IG—with feature-wise occlusion ablation and strict intersection consensus across methods to reduce single-explainer bias. 
We summarize sensitivity and signed effects at raw inputs, at the bridge similarity scaffold, and through the graph convolution, including edge-level sensitivities and targeted edge removals. 
The results show that explainability is most informative when treated as model criticism: it reveals modality dominance, padding and special-token artifacts, dataset-dependent cooperative versus suppressive effects across layers, and chemistry-consistent fragment and composition motifs where methods agree. 
These analyses do not substitute for structural or experimental ground truth, yet they can provide testable hypotheses for downstream validation in computational drug discovery pipelines.
More broadly, applying modern XAI to contemporary DTI/DTA models is still an early pass over the rich structure implicit in trained weights and data—yet even this first layer of scrutiny already helps researchers relate predictions to drug- and target-side representations and to prioritize external validation.
\end{abstract}





\end{frontmatter}



\section{Introduction}

Drug-target interaction (DTI) and drug-target affinity (DTA) prediction models have significantly advanced using deep learning techniques which excel at capturing complex relationships between drugs and proteins. 
However, their inherent complexity often renders them "black boxes", making it challenging to interpret the reasoning behind specific predictions. 
Explainable artificial intelligence (XAI) methods aim to address this challenge by providing transparent insights into model decision-making processes. 

The discrepancy between the vast landscape of DTI/DTA models and the limited application of XAI on such models highlights that many models remain effectively “black boxes” and their decisions are often ignored in biological analysis. 
Consequently, post-hoc \footnote{Pos-hoc methods refer to techniques which relies on auxiliary analytical procedures applied to trained models (after training) in order to infer how predictions are produced.} explainability methods become essential, providing a practical way to extract insights from existing models without requiring re-design, and enabling researchers to trace predictions back to meaningful molecular features, residues, or substructures.

Many drug-target interaction and drug-target affinity prediction models focus on improving performance metrics, often adding layers of complexity without considering how these complexities affect interpretability. 
While such models may claim superior accuracy, high predictive performance alone does not guarantee reliability, generalizability, or applicability in real-world settings. 
Understanding why a model predicts that a particular drug interacts with a specific protein can reveal the underlying mechanisms of binding. 
This interpretability not only validates predictions but also provides valuable insights for drug discovery, uncovering patterns and relationships that may be difficult to identify otherwise \cite{vefghi2025drug}. 
In this way, DTI models can go beyond standard explanations to serve as hypothesis generators, guiding the design of novel therapeutics. 

To our knowledge, this research is one of the first studies in which multiple post-hoc XAI techniques are applied to a black-box (drug-protein interaction) DPI prediction model, aiming to reveal how drugs, proteins, and bridge nodes contribute to model decisions.
Our study emphasizes interpretability rather than model development, highlighting biologically relevant patterns and key predictive features. Our contributions are:

\begin{itemize}
    \item XAI on multi-layer BridgeDPI model: Applied Integrated Gradient (IG), Layer-wise Relevance Propagation (LRP), saliency maps, SmoothGrad, and perturbation-based methods to a multi-layered architecture with high performance metrics.
    \item Bridge node analysis: Explored the role of bridge nodes and edges in linking drug and protein features.
    \item Dataset-level aggregation: Highlighted top scored sections of different input layers in three different datasets using consensus of methods for robustness of results.
\end{itemize}

\section{Related works}

DTI/DTA studies have so far used four families of post-hoc XAI methods that are compatible with current deep learning architectures which are gradient-based attribution, attention visualizations, perturbation and surrogate techniques, and counterfactual edits. 
Other XAI paradigms such as ad-hoc and intristic methods may ultimately prove valuable for DTI but remain largely unexplored in this context.

\subsection{Explainable DTI/DTA models}

Attention-based visualization has become the dominant explainability method in DTI/DTA studies because it reveals important segments by visualizing heatmaps without further processing.
Early attention mechanisms, such as the two-way attention between protein residues and drug atoms proposed by Gao et al. \cite{gao2018interpretable}, established the groundwork for explainable DTI prediction.
Subsequent work, including DeepAffinity \cite{karimi2019deepaffinity}, extended this idea with separate, marginalized, and joint attention modules, enabling interpretability across hierarchical biological contexts.
A further refinement appears in MONN \cite{li2020monn}, which integrates explicit pairwise atom-residue interaction supervision using contact labels from over 13000 protein-ligand complexes.
The ML-DTI model \cite{yang2021ml} enhances interaction awareness through mutual-learning layers connecting drug and protein encoders alongside multi-head and position-aware attention, generating atom- and residue-level saliency distributions.
AffinityVAE \cite{wang2023affinityvae} uses a mutual attention module which computes ligand-to-protein and protein-to-ligand attention weights and an intrinsic Atom-Residue Contact Map module that concatenates protein and ligand features into a two-dimensional interaction map and uses a residual Convolutional network to predict the probability of contact between each residue-atom pair.
A multi-attention fusion strategy follows in MGMA-DTI \cite{li2025mgma}, showcasing attention aggregation at multiple representational scales.
Cross- and self-attention have likewise become central to interpretability.
ArkDTA \cite{gim2023arkdta} leverages ground-truth non-covalent interactions (NCIs) from protein-ligand complexes to form atom-residue interaction matrices regularized by an attention loss term.
ICAN \cite{kurata2022ican} introduces a statistical interpretability assessment aligning attention heatmaps with experimentally confirmed binding sites, whereas AttentionSiteDTI \cite{yazdani2022attentionsitedti} infers self-attention scores over concatenated drug-target embeddings.
A sequential attention cascade emerges in FragXsiteDTI \cite{khodabandeh2024fragxsitedti}, whose learnable latent query attends to pockets, performs self-interaction refinement, and subsequently queries drug fragments.
Expanding on this paradigm, BindingSiteAugmentedDTA \cite{yousefi2023bindingsite} adapts AttentionSiteDTI's graph-based pocket prediction to identify the most relevant binding pockets by reinterpreting protein-ligand complexes as natural language processing-style sequences.
Hierarchical attention frameworks have further advanced explainability: HiGraphDTI \cite{liu2024higraphdti} employs a cross-level hierarchical scheme linking protein segments to atom-, motif-, and global representations, while INGNN-DTI \cite{sun2024ingnn} adopts a nested hierarchical graph architecture to learn molecular and protein embeddings across multiple scales.
Similarly, MultiGranDTI \cite{gong2025multigrandti} exploits learnable assignment matrices that integrate multi-granularity information in a unified graph framework.

Complementing attention mechanisms, gradient-based attribution approaches estimate feature importance by propagating output gradients back to salient input components, revealing model sensitivity to minimal perturbations.
Representative methods include simple saliency maps \cite{simonyan2013deep}, Integrated Gradients \cite{sundararajan2017axiomatic}, Gradient × Input and DeepLIFT-style formulations \cite{shrikumar2017learning}, Grad-CAM \cite{selvaraju2017grad}, Grad-AAM \cite{yang2022mgraphdta}, and GNNExplainer \cite{ying2019gnnexplainer}.
Several adaptations tailor these techniques to biochemical data. Monteiro et al. \cite{monteiro2022explainable} introduced a regression variant of Grad-CAM that produces localization maps for one-dimensional protein sequences and SMILES strings.
GSAML-DTA \cite{liao2022gsaml} incorporates Grad-AAM alongside intra-graph attention weights extracted from protein-encoder GAT layers, while MGraphDTA \cite{yang2022mgraphdta} applies Grad-AAM both to its MGNN backbone and GAT baseline using native attention maps.
MvGraphDTA \cite{zeng2024mvgraphdta} computes input-feature gradients with respect to prediction loss, ranking their magnitudes as per-feature importance scores.
The Structure-Aware GNN \cite{shi2025structure} unifies several feature-attribution approaches—CAM, Grad-CAM, Gradient × Input, and Integrated Gradients—under a single explainability interface, while GNNBlockDTI \cite{deng2025efficient} introduces a streamlined interpretability layer for GNN blocks, inspired by Grad-CAM visualization.

Beyond gradients, perturbation-based methods interpret models by analyzing predictive shifts resulting from controlled input modifications \cite{zintgraf2017visualizing}.
SHAP \cite{lundberg2017unified}, rooted in game-theoretic attribution, extends this idea through local surrogate modeling and Monte-Carlo perturbation sampling.
Its utility has been empirically demonstrated in several DTI frameworks: Ru et al. \cite{ru2023optimization} refined regression-tree feature importance using SHAP values, and VGAN-DTI \cite{kotkondawar2025generative} carried out a feature-level interpretability analysis over engineered fingerprints and physicochemical descriptors using the same principle.
A distinct strand of research explores counterfactual explanations, which seek minimal perturbations capable of altering predicted affinity \cite{wellawatte2022model, schwalbe2024comprehensive}.

Within this paradigm, MACDA \cite{nguyen2021counterfactual} formulates DTI explanation as a multi-agent reinforcement learning task, jointly modifying drug graphs and protein sequences through chemically valid operations.
By optimizing for minimal structural changes that cause maximal shifts in predicted binding affinity, MACDA produces counterfactual drug-protein pairs that illuminate the causal reasoning underlying deep DTA models.

\subsection{BridgeDPI}

Bridge-DPI \cite{wu2022bridgedpi} is a novel drug-protein interaction prediction framework that allows the model to be trained on larger datasets using a smart technique which involves having bridge nodes to allow the model train in batches and update the specific routes form drugs to protein pairs.
Nodes between drug-drug, protein-protein, and drug-protein are not connected directly, therefore bridge nodes become the bottleneck and the main route for drugs and proteins to pass messages between them during the training process.

Bridge nodes importance is mainly metioned, however, no ablation study (i.e. edge weights between bridge nodes and drugs/proteins could be visualized to show the importance of bridge nodes) is done for the bridge nodes to experiment on why they exist in the first place. 
The potential of the bridge nodes in interpretability is unused and the paper goes no far to contribute to the explainability and interpretability of the model.

\section{Materials and Methods}

\begin{figure*}[ht]
  \centering
  \includegraphics[width=\textwidth]{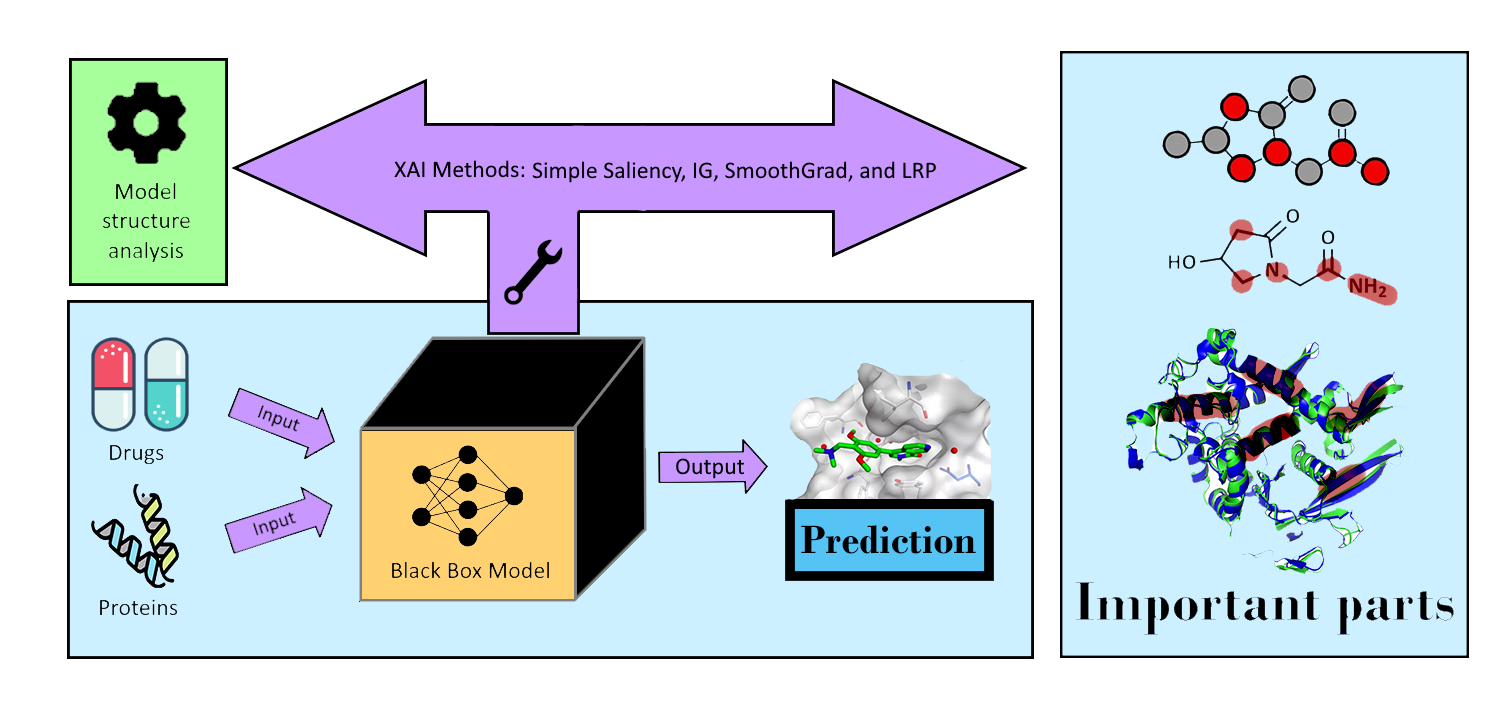}
  \caption{An overview of the main method used in this paper; meaning applying XAI methods on a DPI model to highlight the important parts of the inputs and analyzing various layers of the DPI model.}
  \label{fig:fig1}
\end{figure*}

Multiple XAI methods are used on a black-box model to investigate important parts of the inputs and analyze the structure of the main model as Fig.~\ref{fig:fig1} shows.
BridgeDPI, a multi-layer DPI black-box model is selected as the main model to be analyzed due to its high numbers across different performance measures. 
The analysis demonstrates the potential of the XAI tools used to reveal important regions of the model's input.
Additionally, the modules within BridgeDPI such as bridge nodes are analyzed to understand their specific role in the decision-making process.

\subsection{XAI methods}

A suite of complementary explainability techniques is employed to show the internal reasoning of BridgeDPI model: Simple Saliency, Integrated Gradients (IG), SmoothGrad applied to both IG and Simple Saliency, and Layer-wise Relevance Propagation (LRP). 
Each method provides a distinct perspective on how drug and protein features contribute to the final prediction. 
Combined use of these explainability techniques can mitigate the drawbacks of each method and provide a clear view of models reasoning.

Simple saliency maps \cite{simonyan2013deep} serve as the foundational gradient-based approach for sensitivity analysis. 
By computing the derivative of the output with respect to each input feature, they expose local dependencies between molecular substructures and predicted affinities. 
Saliency maps highlight atoms, residues, and physicochemical descriptors exerting strong influence on model output. 
Although inherently straightforward and computationally efficient, simple saliency maps can display noisy attribution patterns—hence motivating complementary smoothing and integration strategies.

To increase numerical stability and theoretical consistency, Integrated Gradients (IG) \cite{sundararajan2017axiomatic} is adopted, which accumulate gradients along a linear path between a baseline input and the actual representation. 
This integration accounts for the total effect of each feature on the prediction, satisfying sensitivity and implementation invariance axioms. 
IG yields smoother and more reliable interpretations by identifying features contributing most strongly to binding affinity.

SmoothGrad \cite{smilkov2017smoothgrad} can be applied on IG to mitigate the noise inherent in single-pass gradient estimation. 
SmoothGrad-IG averages multiple IG computations over stochastically perturbed inputs, generating visually coherent attribution maps that preserve underlying causal relationships. 
A parallel smoothing strategy can also be integrated with Simple Saliency. 
SmoothGrad-Saliency averages raw gradient outputs over perturbed samples, reducing background noise and amplifying stable regions. 

Layer-wise Relevance Propagation (LRP) \cite{bach2015pixel} can be incorporated to complement gradient-based analyses. 
Unlike differential sensitivity methods, LRP redistributes the model's output relevance backward through the network layers, quantitatively decomposing predictions into neuron-level contributions. 
This allows tracing predicted affinity scores directly to input features while conserving total relevance across the architecture.

\subsection{Datasets}

Three different datasets are used in this study including BindingDB, C.elegans and Human datasets.

BindingDB dataset: The affinity data of 2286319 drug-protein pairs from corresponding research papers is collected,
where 8536 proteins and 989383 drugs are included \cite{gilson2016bindingdb} \footnote{\url{https://www.bindingdb.org/bind/index.jsp}}. Most drug-target pairs in BindingDB are reported as $IC50$ measurements, followed by $K_i$, $EC50$, and $K_d$ values.
Gao version \cite{gao2018interpretable} \footnote{\url{https://github.com/IBM/InterpretableDTIP}} filters the data based on $IC50$ values and convert the affinity scores to binary interactions using thresholds of 100 $nM$ and 10000 $nM$.

C.elegans and Human datasets: These datasets are constructed by combining a set of highly credible negative drug-protein samples via an in silico screening method with the known positive samples \cite{liu2015improving}. 
The balanced versions of these datasets are used here \cite{tsubaki2019compound} \footnote{\url{https://github.com/masashitsubaki/CPI_prediction}}. 
The C.elegans dataset has 7786 drug-protein pairs. 
Proteins in this dataset are from the species Caenorhabditis elegans, a small nematode worm widely used as a model organism in biology. 
C.elegans proteins include homologs of human proteins but may also include nematode-specific proteins.
The Human dataset has 6728 drug-protein pairs in total. 
Proteins in this dataset are human proteins, often from a wide range of classes such as Enzymes (e.g., kinases like EGFR, CDK2), Receptors (GPCRs, nuclear hormone receptors), Ion channels.

In general, Models should perform better on Human and C.elegans datasets due to high overlap in training and test entities, whereas Gao's dataset allows more generalization to unseen compounds and proteins. 

\begin{table*}[ht]
\centering
\tiny
\caption{Datasets Statistics. For the Gao dataset in first part of the table, the values in parentheses are totals for all cataloged proteins/drugs (some may not appear in any interaction), while the values before the parentheses count only those that actually participate in interactions.}
\begin{tabular*}{\textwidth}{@{\extracolsep{\fill}} lccc}
\toprule
Features/Datasets & Gao & C.elegans & Human\\
\midrule
\textbf{\# Unique Proteins in interactions} & 813 (842) & 1876 & 2001\\
\textbf{\# Unique Drugs in interactions} & 49752 (171926) & 1767 & 2726\\
\textbf{\# Total Amino Acids} & 471022 (489412) & 1074521 & 1181424\\
\textbf{\# Total Drug Atoms} & 5165804 (1511956) & 32915 & 61577\\
\textbf{\# Unique Amino Acids} & 21 & 21 & 22\\
\textbf{\# Unique Drug Atoms} & 21 (26) & 28 & 62\\
\textbf{Average Protein sequence Length} & 579.36 (581.24) & 572.77 & 590.41\\
\textbf{Average Drug sequence Length} & 30.38 (30.04) & 18.62 & 22.58\\
\midrule
\textbf{\# Train interactions} & 50155 (28240 pos / 21915 neg) & 6228 (3139 pos / 3089 neg) & 5382 (2700 pos / 2682 neg)\\
\textbf{\# Valid interactions} & 5607 (2831 pos / 2776 neg) & N/A & N/A\\
\textbf{\# Test interactions} & 5508 (2706 pos / 2802 neg) & 1558 (754 pos / 804 neg) & 1346 (664 pos / 682 neg)\\
\textbf{\% Protein warm setting} & 90.3 \% valid / 91.0 \% test & 85.5 \% valid & 81.8 \% valid\\
\textbf{\% Drug warm setting} & 30.1 \% valid / 30.3 \% test & 74.8 \% valid & 60.7 \% valid\\
\midrule
\textbf{Protein duplicacy index} & 75.36 & 4.15 & 3.36\\
\textbf{Drug duplicacy index} & 1.23 & 4.41 & 2.47\\
\textbf{\% Proteins with >=2 interaction} & 83.9\% & 69.0\% & 61.9\%\\
\textbf{\% Drugs with >=2 interaction} & 15.5\% & 54.9\% & 38.8\%\\
\textbf{\# Max edges/protein} & 1459 & 504 & 425\\
\textbf{\# Max edges/drug} & 72 & 399 & 264\\
\midrule
\textbf{Aliphatic (A, V, L, I, M)} & 30.52\% & 30.58\% & 30.78\%\\
\textbf{Polar uncharged (S, T, N, Q)} & 21.04\% & 21.65\% & 20.85\%\\
\textbf{Positive (K, R, H)} & 13.65\% & 14.10\% & 13.38\%\\
\textbf{Aromatic (F, Y, W)} & 8.43\% & 8.22\% & 8.54\%\\
\textbf{Sulfur (C, M)} & 4.68\% & 4.25\% & 4.54\%\\
\textbf{Special (G, P)} & 12.4\% & 10.96\% & 12.58\%\\
\textbf{Acidic residues (D, E)} & 11.64\% & 12.80\% & 11.70\%\\
\textbf{P with any aromatic} & 100\% & 100\% & 100\%\\
\textbf{P with >=2 Cys} & 96.8\% & 90.4\% & 94.7\%\\
\midrule
\textbf{Any rings / Aromatic rings} & 99.7\% / 98.6\% & 72.0\% / 61.5\% & 73.8\% / 52.8\%\\
\textbf{Heterocycles} & 92.9\% & 50.1\% & 50.3\%\\
\textbf{Fused systems} & 68.1\% & 37.1\% & 41.2\%\\
\textbf{Avg rings/molecule} & 3.98 & 1.95 & 2.33\\
\textbf{Avg aromatic rings} & 2.95 & 1.28 & 1.13\\
\textbf{Top Murcko scaffold} & benzene (c1ccccc1, 1.04\%) & benzene (c1ccccc1, 10.07\%) & benzene (c1ccccc1, 6.05\%)\\
\bottomrule
\end{tabular*}
\label{tab:tab1}
\end{table*} 

\begin{figure*}[ht]
  \centering
  \includegraphics[width=\textwidth]{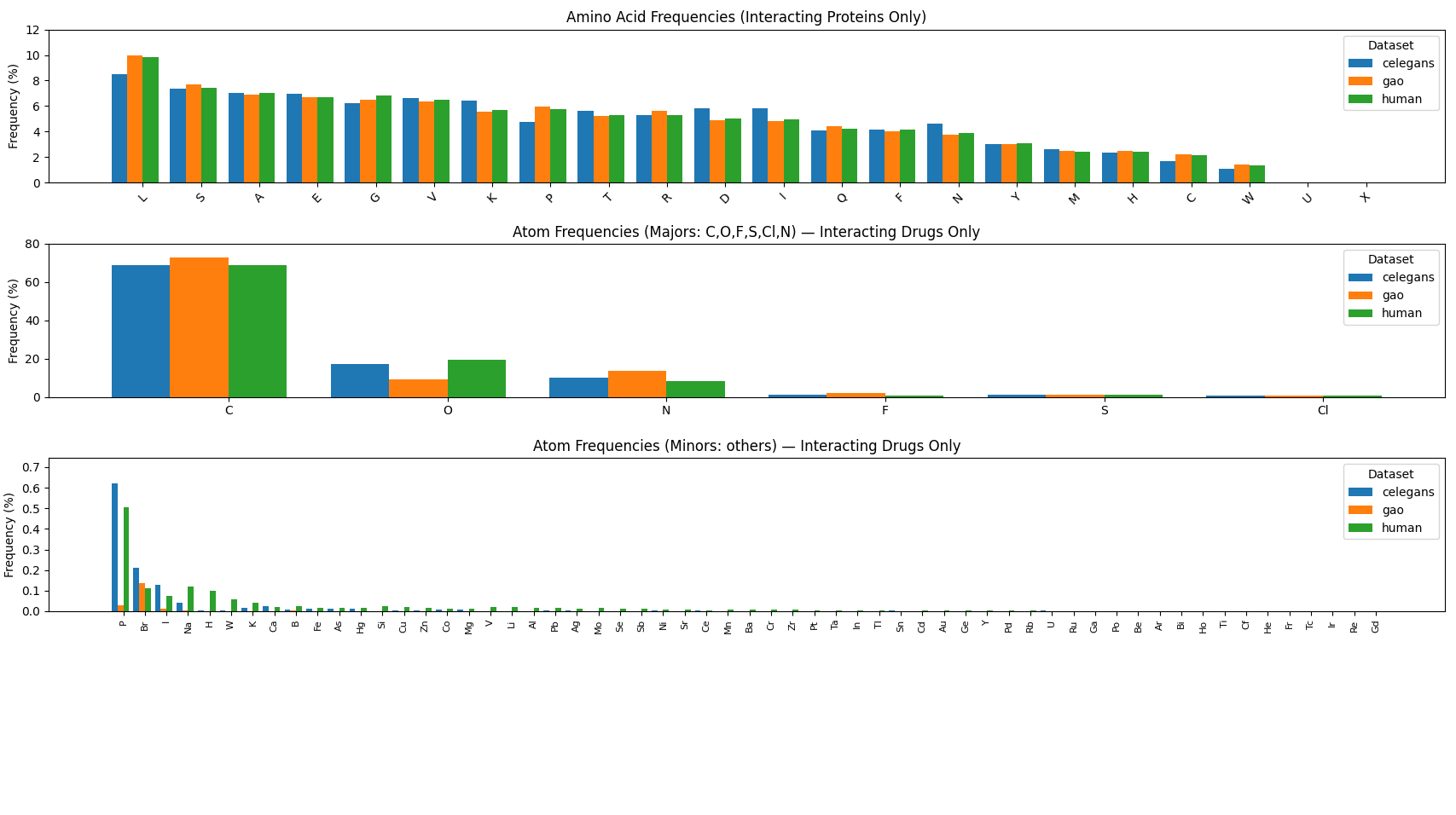}
  \caption{Amino Acid and Atom frequencies. C.elegans AA vocab (L S A E V K G D I T R P N F Q Y M H C W U), Gao AA vocab (L S A E G V P R K T D I Q F N Y H M C W U), Human AA vocab (L S A G E V P K T R D I Q F N Y H M C W U X), C.elegans Atom vocab: (C O N F S Cl P Br I Na Ca K Fe As Hg Mg B Co Zn Ce Pb U Ni Cu Sn Ag H W), Gao atom vocab: (C N O F S Cl Br P I Na B Si Se Re V Ru K Fe As Gd Sb), Human atom vocab: (C O N S F Cl P Na Br H I W K B Si Cu V Ca Li Zn Fe As Al Pb Hg Mo Ag Se Co Mg Sb Sr Mn Ni Ba Cr Zr Pt Tl Ta In Cd Ge Rb Ce Y Pd Au Sn Ho Ru Cf Ar Ga Be Po Bi Ir Ti Fr Tc He). Bars which cannot be seen are not zero, they are very low.}
  \label{fig:fig2}
\end{figure*}

Table~\ref{tab:tab1} summarizes the three interaction corpora used for training and evaluation. 
Collectively, they span markedly different scales, reuse patterns, and small-molecule complexity, which jointly constrain both predictive difficulty and how interpretability results should be read.

The Gao collection couples a comparatively small protein inventory that participates in interactions ($\approx800$ proteins in pairs; larger totals exist when catalog entries never observed in pairs are included) with an exceptionally large compound inventory ($\approx5\times10^4$ drugs in pairs versus $\approx1.8-2.7\times10^{3}$ on C.elegans and Human). 
Consequently, aggregated amino-acid and atom counts are dominated by Gao's breadth of chemistry even though average protein lengths remain similar across datasets ($\approx573-590$ residues). 
Average molecular size along the modeled drug representation is largest on Gao ($\approx30$ atoms) versus shorter typical graphs on C.elegans ($\approx19$) and Human ($\approx23$), increasing representation complexity and sparsity risk for attribution maps on the screening-style corpus.

Training splits are approximately balanced between positives and negatives where reported; Gao additionally includes dedicated validation partitions. 
Warmth differs sharply by modality on Gao: protein-side warmth is high ($\approx90\%$ validation/test), whereas drug-side warmth is low ($\approx30\%$), so evaluation emphasizes novel compounds against recurrent targets—a stringent cold-drug regime. 
C.elegans and Human instead show moderate protein warmth ($\approx82-86\%$ validation where listed) paired with substantially higher drug warmth ($\approx61-75\%$), shifting failure modes toward protein novelty (relative to Gao's compound novelty) rather than uniformly identical evaluation pressures across benchmarks.

Protein duplicacy is extreme on Gao (index $\approx75$ versus $\approx3-4$ on the other corpora): many interactions reuse the same proteins ($\approx84\%$ with$\ge$2 edges; maxima exceeding $10^3$ partners per protein), whereas drug duplicacy remains modest ($\approx1.2$; $\approx16\%$ of drugs with $\ge$2 interactions). 
C.elegans and Human exhibit more balanced reuse, including higher fractions of drugs appearing in multiple pairs. 
For modeling and explanation, hub structure implies strong protein-centric averaging on Gao—gradient or occlusion signals may reflect targets seen under many chemotypes—whereas organism subsets emphasize different reuse geometries when interpreting consensus features.

Amino-acid inventories are stable (21-22 residue types with similar coarse bins; $\approx31\%$ aliphatic across sets). 
Thus differences in protein-level explanations across benchmarks are unlikely to reflect incompatible alphabets; differences more plausibly reflect task geometry, warmth, and organism-specific sequence composition. 
Fig.~\ref{fig:fig2} corroborates shared marginal residue frequencies dominated by abundant residues (L, S, A, G, V, acidic/basic types also frequent). 
These histograms characterize whole-sequence corpus composition, so they should not be equated with binding-site enrichment unless complemented by positional or structural analyses

Drug topology diverges strongly: Gao overwhelmingly contain rings and aromatic systems with abundant heterocycles/fused frameworks (several rings per molecule on average), whereas C.elegans and Human show lower aromatic-ring incidence and fewer rings per molecule—consistent with less uniformly drug-like libraries and differing scaffold diversity (benzene-type Murcko cores rank first in each set but at dataset-dependent prevalence). 
Fig.~\ref{fig:fig2} elemental distributions show C/O/N dominance as expected; halogens and heteroatoms appear at smaller marginal rates with dataset-specific shifts (interpret cautiously because preprocessing, protonation assumptions, and graph featurization strongly shape counted atom types—especially hydrogens and alkali ions when present).

Cross-dataset explanation comparisons should be read jointly with cold-drug severity on Gao, hubbed proteins, mean molecular complexity, and topological priors: identical attribution machinery can emphasize different apparent pharmacophores simply because the underlying chemistry frequency tables differ.

\subsection{Training of the main model}

As Table~\ref{tab:tab2} shows, the model demonstrates strong performance with high accuracy and excellent discriminative ability, indicating that it has apparently learned meaningful patterns in the drug-target interaction data. 
The balanced precision and recall suggest the model captures both positive and negative interactions effectively, while the low loss values indicate confident predictions. 
The reasonable gap between training and validation performance shows good generalization without overfitting, meaning the model has learned generalizable features rather than memorizing the training data. 
The high discriminative ability particularly indicates that the model has identified robust patterns that should manifest as strong, interpretable gradients when analyzing which protein and drug features contribute most to the interaction predictions.

Although biological validation of explainability results (e.g., comparison of model-attributed residues or atoms with experimentally confirmed binding pockets) would provide the most direct evidence of interpretability, such validation is not yet feasible at scale. 
The commonly used datasets—BindingDB, Davis/KIBA, and C.elegans/Human—rarely include residue-level structural annotations or complete co-crystal data. 
Even when structural information is available in external databases such as PDB or PDBBind, only a small subset of targets overlap with those used in benchmark datasets, making systematic mapping unreliable. 
Therefore, this study focuses on input-level interpretability—identifying the regions of drugs and proteins that most strongly influence model predictions—while recognizing that deeper biological alignment remains an open challenge for the field.

The model architecture was modified to replace the element-wise multiplication operation between protein and drug components with concatenation to facilitate more interpretable gradient analysis. 
This architectural change was motivated by the need to obtain cleaner, independent gradients for each modality during backpropagation. 
While multiplication creates coupled gradients where changes in one component affect the gradient of the other, concatenation preserves independent gradient flow, allowing for more precise attribution analysis.
The transition from multiplication to concatenation in the final layer, while beneficial for independent gradient analysis of protein and drug components, inadvertently diminished the interpretability of bridge nodes by removing their role as mediators of protein-drug interactions.

In gradient methods, both input $\times$ gradient and raw gradient are used, but the results were very similar, therefore, raw gradient is used in all sections.

\begin{table*}[htbp]
\centering
\tiny
\caption{Performance metrics (mean ± std) over 5 runs for testing bridge nodes on three datasets.}
\begin{tabular*}{\textwidth}{@{\extracolsep{\fill}} lccccccc}
\toprule
\textbf{ } & \textbf{Accuracy} & \textbf{AUC} & \textbf{Precision} & \textbf{Recall} & \textbf{F1}\\
\midrule
Gao w/ Bridge & \textbf{0.9180 ± 0.0014} & \textbf{0.9700 ± 0.0023} & 0.9316 ± 0.0186 & \textbf{0.8998 ± 0.0223} & \textbf{0.9152 ± 0.0033}\\
Gao w/o Bridge & 0.9062 ± 0.0082 & 0.9680 ± 0.0007 & \textbf{0.9348 ± 0.0193} & 0.8704 ± 0.0362 & 0.9008 ± 0.0113\\
\midrule
C.elegans w/ Bridge & 0.9730 ± 0.0062 & 0.9958 ± 0.0016 & 0.9704 ± 0.0077 & \textbf{0.9744 ± 0.0078} & 0.9724 ± 0.0067\\
C.elegans w/o Bridge & \textbf{0.9736 ± 0.0076} & \textbf{0.9962 ± 0.0013} & \textbf{0.9720 ± 0.0096} & 0.9736 ± 0.0102 & \textbf{0.9728 ± 0.0079}\\
\midrule
Human w/ Bridge & \textbf{0.9638 ± 0.0033} & 0.9926 ± 0.0023 & \textbf{0.9728 ± 0.0057} & 0.9530 ± 0.0114 & \textbf{0.9626 ± 0.0039}\\
Human w/o Bridge & 0.9606 ± 0.0062 & \textbf{0.9928 ± 0.0019} & 0.9550 ± 0.0133 & \textbf{0.9664 ± 0.0052} & 0.9606 ± 0.0060\\
\bottomrule
\end{tabular*}
\label{tab:tab2}
\end{table*}

\begin{figure}[ht]
  \centering
  \includegraphics[width=\columnwidth]{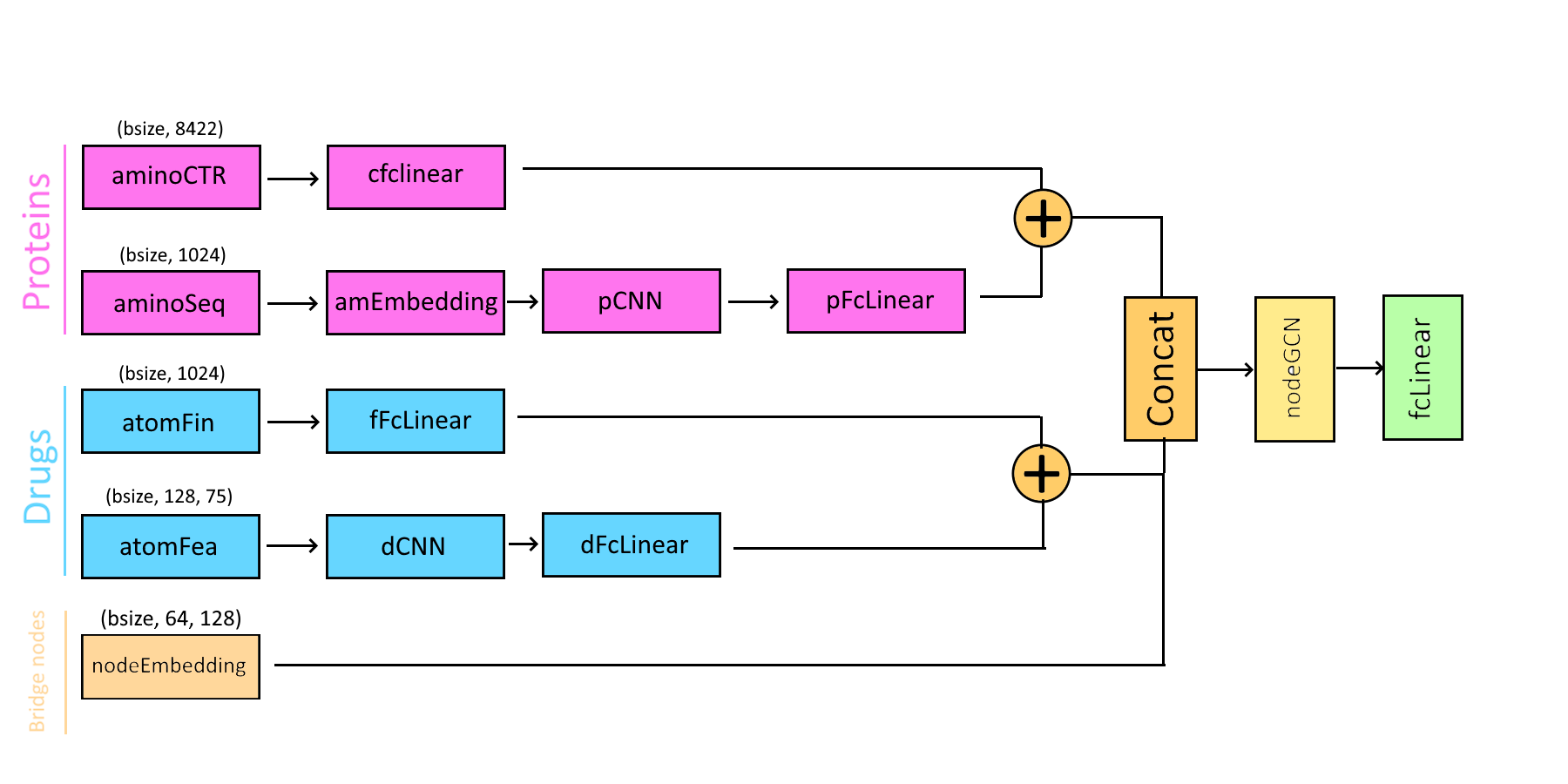}
  \caption{Overview of the BridgeDPI pipeline including inputs and different modules.}
  \label{fig:fig3}
\end{figure}

Preprocessing is done exactly like the preprocessing that is implemented in BridgeDPI paper \footnote{\url{https://github.com/enai4bio/BridgeDPI}}. Max sequence length of proteins and drugs are set at 1024 and 128, respectively.
As Fig.~\ref{fig:fig3} shows,BridgeDPI as a complex model contains four inputs including:
\begin{itemize}
\item AminoSeq: primary sequence vectors of proteins, 
\item AminoCTR: concatanation of 1-mer,2-mer, and 3-mer tensors of proteins, 
\item AtomFin: Morgan fingerprint vectors of drugs acquired from SMILES of drugs, 
 \item AtomFea: graph features from SMILES for drugs.
\end{itemize}

\section{Experiments and Results}

All experiments are conducted using default parameters of BridgeDPI paper which some of them are mentioned in Table~\ref{tab:tab3}. If any change in parameters is done, it is mentioned in the corresponding section.
For gao dataset, training split is unique across runs, but for human and C.elegans preprocess is performed 5 times to get different splits as it is random split.
Main performance measures may be different than the number that is reported in the original paper due to the differences in parameters and fine-tuning.
Test set of Datasets are used for the post-hoc explanations. 
The Selected method for sampling is Stratified Sampling and procedures are repeated five times to get the final mean result.
Furthermore, batches and minibatches across runs are also used and average vectors between them are reported.

Unlike other studies in which they test several case studies to prove the alignment of XAI output (area attribution scores) and biological explanations (known pockets) by comparing them, in this work scores are aggregated over the test set to find out the overall importance of the inputs for the model.

\begin{table}[ht]
  \centering
  \caption{Hyperparameters of trained BridgeDPI model. pCNN: protein TextCNN, pFcLinear: drug FFN, cFcLinear: drug FFN, dCNN:drug TextCNN, dFcLinear: drug FFN, fFcLinear: drug FFN}
  \label{tab:tab3}
  \begin{tabular}{cc}
    \toprule
    FeatureSize of pCNN & 24\\
    Filter size of pCNN & 25\\
    Filter num of pCNN & 64\\
    Neuron num of pFcLinear & 128\\
    Neuron num of cFcLinear & 1024, 128\\
    
    Filter size of dCNN & 7\\
    Filter num of dCNN & 64\\
    Neuron num of dFcLinear & 128\\
    Neuron num of fFcLinear & 1024, 256, 128\\
    
    Num of bridge nodes & 64\\
    Neuron num of GNN & 128, 128, 128\\
    Decoder neuron num & 128, 1\\
    Optimizer & Adam (lr:0.001)\\
    Weight decay & 0.001\\
    batch size & 512\\
    Number of epochs & 100\\
    Patience & 30\\
    Dropouts & 0.5\\
    
    \bottomrule
  \end{tabular}
\end{table}


In general, layerwise gradient analysis is used to investigate the different sections of the main model.
According to bridgeDPI pipeline in Fig.~\ref{fig:fig3}, layers which their analysis could have important results are gcn-output, gcn-input, GNN module, and the input layer.
Other layers are also analyzed and have similar results to these four sections, therefore these results are not metioned further.

\subsection{Bridge node importance}

This section discusses the importance of the GNN module in BridgeDPI framework.

As Table~\ref{tab:tab2} shows model is trained both with bridge nodes and without bridge nodes to assess the effectiveness of the graph structure and graph neural network in the final prediction. 
As the results show, the drop in performance measures are so little that it indicates the low contribution of bridge nodes to the prediction overally.

\begin{figure}[ht]
  \centering
  \includegraphics[width=\columnwidth]{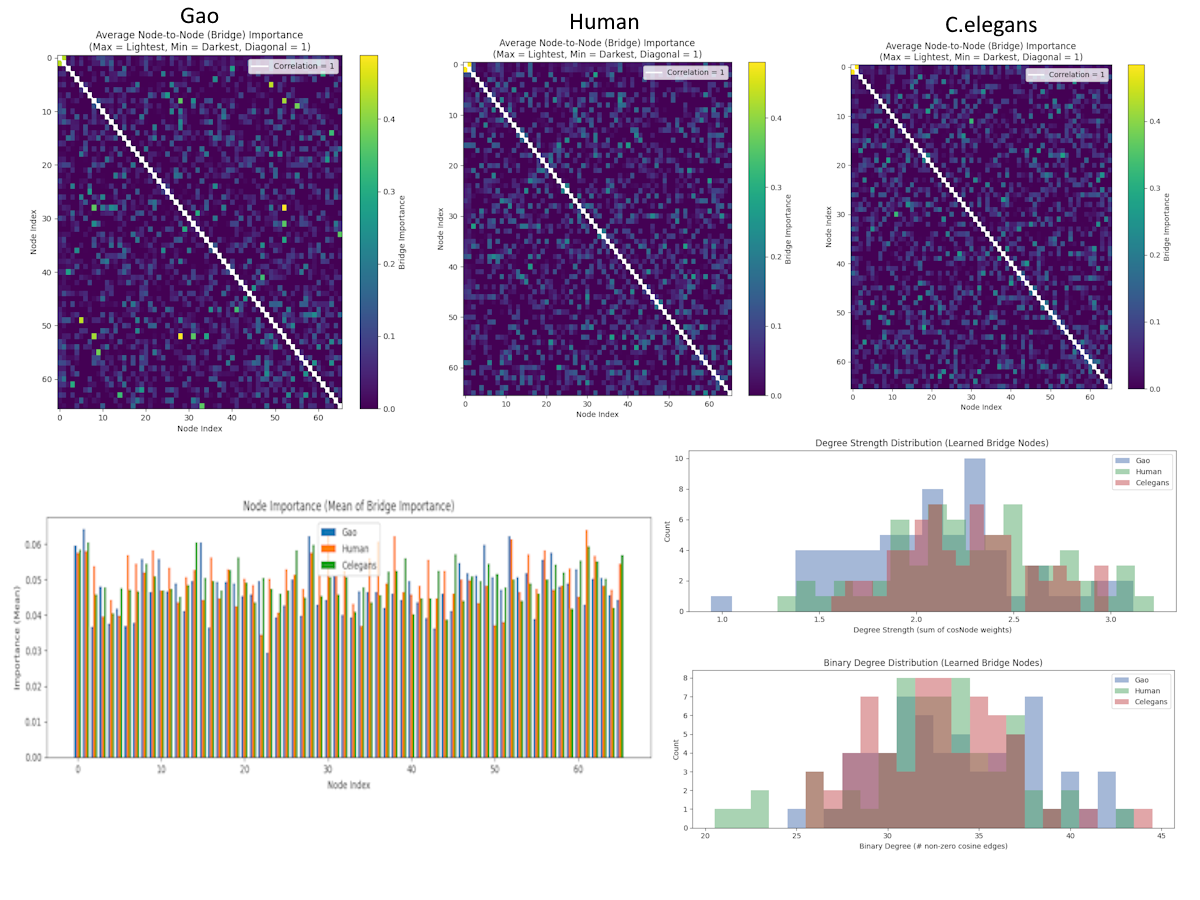}
  \caption{Plots of first analysis in bridge node importance.}
  \label{fig:fig4}
\end{figure}

As Fig.~\ref{fig:fig4} shows, the heatmaps summarize the batch-averaged rectified cosine similarity matrix $\overline{\mathbf{C}}$ between pre-GCN nodes: protein (index~0), drug (index~1), and learned bridge rows ($\geq 2$).
Entries are cosine similarities with $\ell_2$-normalized node vectors, negative values removed, and the diagonal set to~$1$; the figure masks the diagonal for color scaling and draws it separately, so off-diagonal color reflects inter-node coupling only.
Across Gao, Human, and C.elegans, off-diagonal mass is mostly low with scattered brighter pairs, indicating weak average coupling for most node pairs, a few stronger pairs, and no clear block or community structure; the overall sparsity pattern is qualitatively similar between corpora.

The grouped bars plot node-level summaries of the same matrix: the row sum and row mean of $\overline{\mathbf{C}}$ over all node indices.
Values are comparatively uniform, with only modest peaks and troughs, which suggests that aggregate cosine mass is spread across coordinates rather than concentrated on a tiny hub subset, and that dataset-to-dataset shifts in node ranking are mild at this summary level.

Connectivity histograms refer only to learned bridge nodes ($\geq 2$).
Degree strength is the row sum of $\overline{\mathbf{C}}$ with the diagonal zeroed (sum of off-diagonal rectified cosines).
Binary degree counts off-diagonal entries that are strictly positive (no extra threshold).
Distributions overlap substantially; Human shows a slightly heavier right tail in degree strength, while Gao is shifted toward somewhat larger binary degrees in the upper range, with Human and C.elegans more central.
Overall, the bridge acts as a fairly dense rectified-cosine scaffold with small dataset-specific shifts in edge mass and sparsity rather than a qualitative change in layout.

\begin{figure}[ht]
  \centering
  \includegraphics[width=\columnwidth]{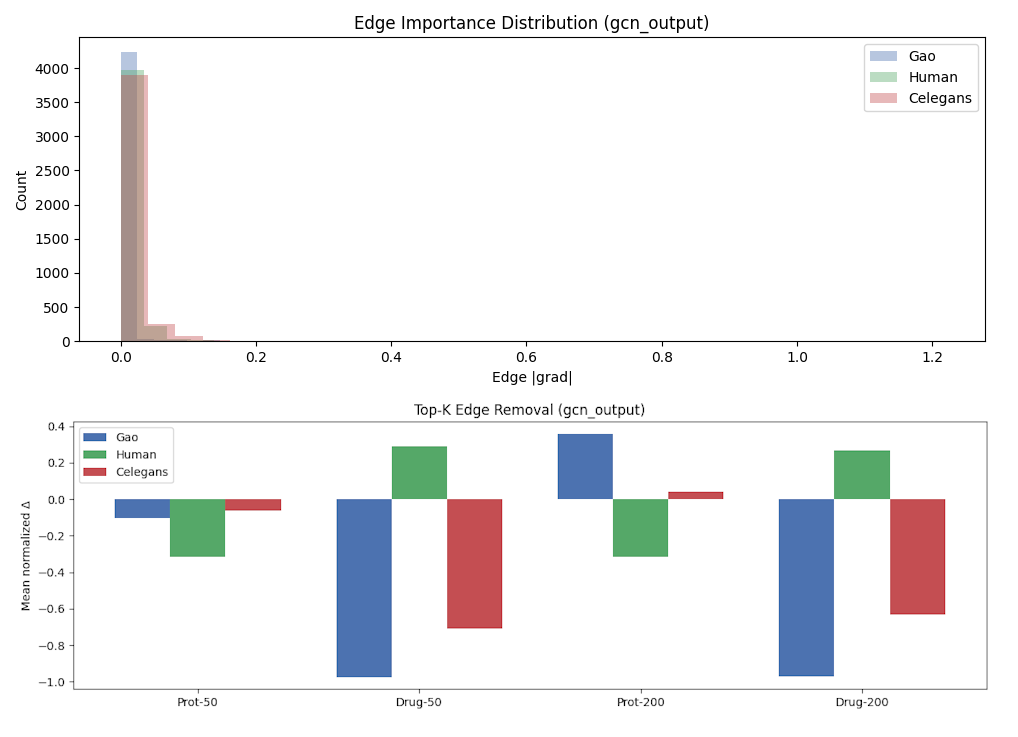}
  \caption{Plots of Second analysis in bridge node importance.}
  \label{fig:fig5}
\end{figure}

To relate these similarities to predictive behavior at the GCN, we analyze edges in the normalized adjacency actually used in message passing (Fig.~\ref{fig:fig5}).
Pre-GCN node features are fixed (detached); the trainable part enters through an adjacency tensor $\mathbf{A}$ obtained from the same rectified cosines with ones on the diagonal.
The GCN uses $\mathbf{pL}=\mathbf{D}^{-1/2}\mathbf{A}\mathbf{D}^{-1/2}$ with $D_{ii}=\sum_j A_{ij}$, followed by the graph convolution and final regressor yielding a scalar score~$y$ per sample.

Edge-level sensitivity scores are computed by treating $\mathbf{A}$ as a differentiable variable.
Unless noted otherwise, our implementation uses integrated gradients along straight-line paths from an identity adjacency to $\mathbf{A}$ (25 steps), accumulated for up to $\min(B,128)$ samples, averaged, and with the diagonal zeroed before visualization; alternatively, plain $|\partial y/\partial \mathbf{A}|$ can be used by changing the attribution mode.
The overlaid histograms show the distribution of these scores over off-diagonal entries (heavy tail: most edges near zero, a sparse subset much larger).

For top-$K$ interventions ($K\in\{50,200\}$), undirected off-diagonal pairs are ranked separately for protein-bridge edges (incident to node~0) and drug-bridge edges (incident to node~1) using the same sensitivity scores.
For each family, the $K$ highest-ranked entries of $\mathbf{A}$ are set to zero symmetrically, the diagonal is restored to~$1$, $\mathbf{pL}$ is rebuilt, and $y$ is recomputed without gradients.
We report the batch mean of $(y_{\mathrm{base}}-y_{\mathrm{abl}})/(|y_{\mathrm{base}}|+10^{-6})$.
Positive values mean removal tends to decrease the score (net supportive under this perturbation); negative values mean removal tends to increase it (net suppressive).

The histograms are strongly heavy-tailed on all three datasets, with Gao showing slightly more mass at small non-zero sensitivities; the top-$K$ bars add directionality.
For $K{=}50$, protein-bridge removals are net suppressive on Gao and especially Human, and weakly suppressive on C.elegans; drug-bridge removals are strongly suppressive on Gao and C.elegans but net supportive on Human.
For $K{=}200$, Gao protein-bridge removals become net supportive while drug-bridge removals stay suppressive; Human protein-bridge removals stay suppressive; on C.elegans, drug-bridge removals stay suppressive and protein-bridge removals become weakly supportive.

Thus predictive sensitivity is concentrated on a small edge subset, while the net supportive versus suppressive role of the highest-sensitivity protein- versus drug-bridge connections is dataset-dependent; sign changes when increasing $K$ are expected because larger ablations interact nonlinearly through renormalization and the GCN stack.

Overall, at the larger $K$ setting illustrated in Fig.~\ref{fig:fig5}, Gao is broadly consistent with net supportive protein-bridge mass together with net suppressive drug-bridge mass, Human with the opposite pattern, and C.elegans with weakly supportive protein-bridge mass together with net suppressive drug-bridge mass.

Overall, the aggregate bridge visualizations suggest a fairly uniform rectified-cosine scaffold rather than a sharply modular or community-structured graph; correspondingly, the static node-to-node similarity maps alone carry limited discriminative story compared with the heavy-tailed task-dependent edge sensitivities at the GCN input.
This does not diminish the architectural rationale of Bridge-DTI: the contribution is largely operational-a batchable construction in which protein, drug, and learned bridge embeddings jointly induce a dense similarity matrix that can be symmetrized and fed through a GCN each step, enabling training on large interaction corpora without building an explicit pairwise graph per mini-batch in an ad-hoc way.
Conceptually, using rectified cosine affinities as nonnegative edge weights before normalized convolution resembles similarity-gated message passing; however, it should not be equated with a general graph attention network (GAT), which typically learns attention coefficients with a distinct parameterization and normalization scheme.

\subsection{GCN input and output layers}

Following sections discuss the analysis of GCN layer input and output.

\subsubsection{GCN output}

\begin{figure}[ht]
  \centering
  \includegraphics[width=\columnwidth]{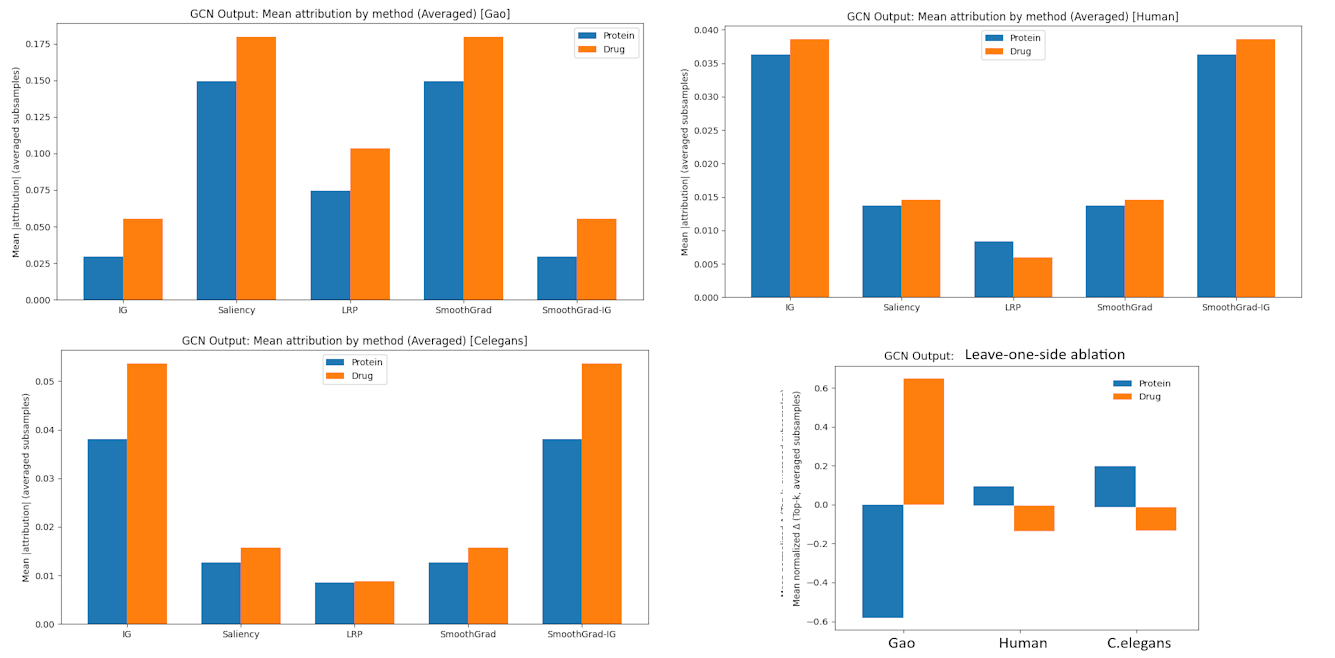}
  \caption{GCN output result plots.}
  \label{fig:fig6}
\end{figure}

Analyzing the GCN output is useful because it characterizes how protein- and drug-side latent features are shaped by message passing before the final encoder produces the prediction.
Comparing these summaries to explanations at the GCN input (and to the bridge similarity structure) helps separate what the graph module does to each modality from what the raw encoders provide.

We quantify the two sides with complementary summaries, each averaged over multiple random stratified subsamples for stability.
First, we report mean absolute attributions for the 128-dimensional protein and drug latents using IG, plain saliency, LRP, SmoothGrad, and SmoothGrad-IG, which summarize local sensitivity magnitude to perturbations on each side.
Second, we perform top-$k$ ablation by zeroing the highest-ranked latent dimensions on each side (separately) and measuring the normalized change in the prediction; the sign separates net supportive effects (positive $\Delta$) from net suppressive effects (negative $\Delta$).
Jointly, magnitude and signed ablation distinguish "what the model reacts to strongly" from "what tends to push the score up versus down" under coordinated removal.
When $k$ equals the full latent width (128), the procedure reduces to leave-one-side ablation of that modality at the GCN output.

Mean $|\text{attribution}|$ can be misleading in isolation if opposing influences cancel in the attributions; pairing it with signed top-$k$/leave-one-side ablation reduces that ambiguity.
These analyses are nonetheless layer-local: they do not trace compensatory rerouting in earlier layers, nor do they remove dataset composition effects (e.g., cold-drug evaluation regimes).

\paragraph{Results (Fig.~\ref{fig:fig6})}
\textbf{Gao.} Across methods, mean $|\text{attribution}|$ is higher for drug latents than for protein latents, indicating greater local sensitivity to perturbations on the drug-side GCN output.
The leave-one-side ablation is directionally consistent with a drug-driven positive contribution: drug-side removal yields a positive normalized $\Delta$ (supportive latents), whereas protein-side removal yields a negative $\Delta$ (net suppressive latents at this layer under joint top-$k$ selection).

\textbf{Human.} For IG, saliency, SmoothGrad, and SmoothGrad-IG, mean $|\text{attribution}|$ is modestly larger on the drug side; LRP is the exception, assigning larger magnitude to the protein side.
Ablation nonetheless reverses the directional emphasis relative to Gao: protein-side removal produces a positive $\Delta$, while drug-side removal produces a negative $\Delta$.
Thus, Human is a case where gradient-style magnitudes need not align with the net push/pull inferred from coordinated removal: the protein GCN output appears more supportive for the score, whereas the selected highly ranked drug latents behave more suppressively at this layer.

\textbf{C.elegans.} Mean $|\text{attribution}|$ is consistently larger on the drug side across methods (the gap is especially visible for IG and SmoothGrad-IG), so perturbations of drug latents produce the strongest local attributions.
However, leave-one-side ablation does not follow the same ranking: protein-side removal yields a positive $\Delta$ (supportive), whereas drug-side removal yields a negative $\Delta$ (suppresssive).
In other words, the worm model can be locally most sensitive to drug coordinates while the coordinated top-$k$ mass on the drug side acts, on average, as a brake on the score; the protein side provides the net supportive offset in this ablation view.

\subsubsection{GCN input}

\begin{figure}[ht]
  \centering
  \includegraphics[width=\columnwidth]{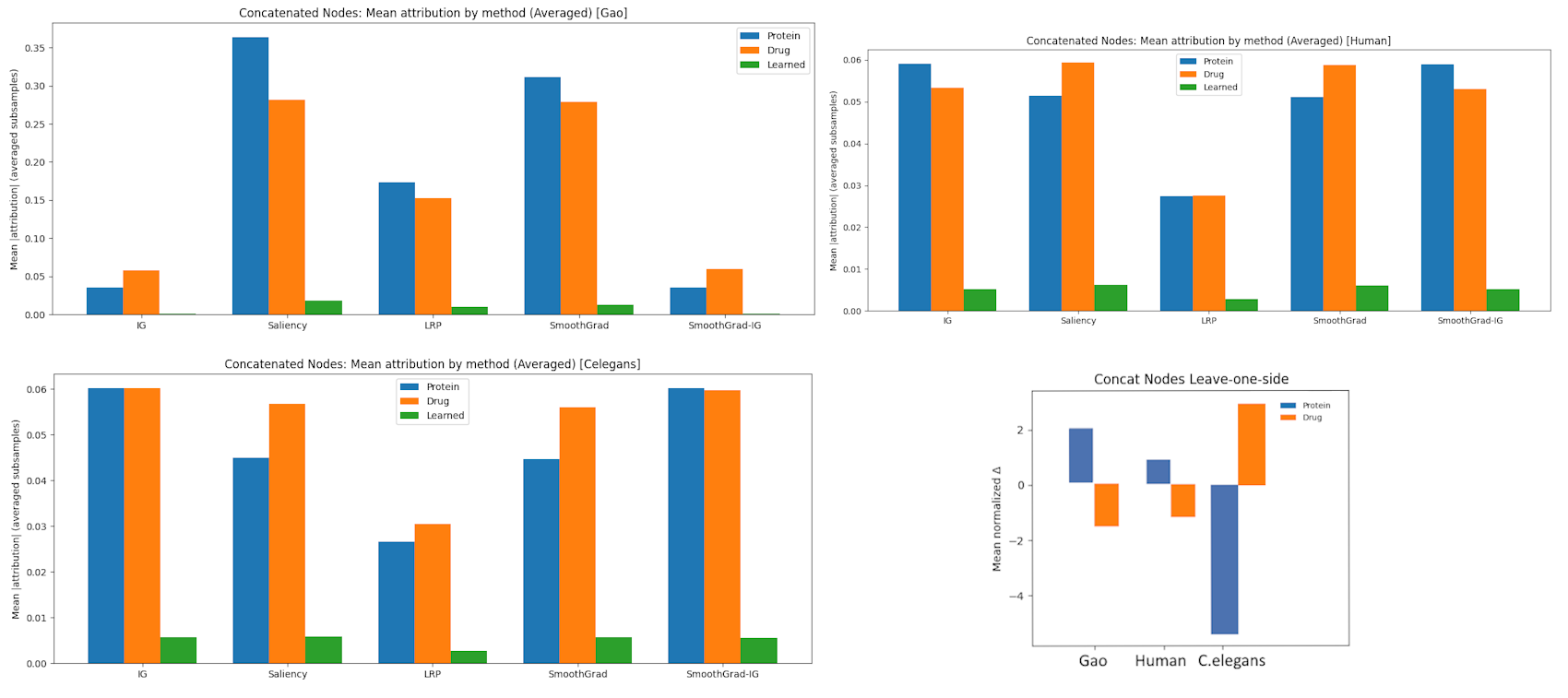}
  \caption{Explanations at the GCN input (pre-message-passing node tensor). \textbf{Top:} mean absolute attributions (IG, saliency, LRP, SmoothGrad, SmoothGrad-IG), averaged over stratified subsamples, for the three concatenated node groups-protein (node~0 embedding), drug (node~1 embedding), and learned bridge rows. \textbf{Bottom:} leave-one-side ablation on the concatenated nodes: mean normalized $\Delta=(y_{\mathrm{base}}-y_{\mathrm{abl}})/(|y_{\mathrm{base}}|+\varepsilon)$ when an entire side is removed. Positive $\Delta$ indicates that side is net supportive (removal lowers the score); negative $\Delta$ indicates net suppressive (removal raises the score).}
  \label{fig:fig7}
\end{figure}

The GCN input is the tensor formed by concatenating protein-side, drug-side, and learned bridge embeddings into a single node matrix before cosine-based adjacency construction and graph convolution.
This layer is a natural checkpoint because it separates (i)~how much gradient-based attributions target each source of features from (ii)~how much coordinated removal of an entire modality changes the prediction.

We summarize the same five explanation methods used elsewhere (IG, saliency, LRP, SmoothGrad, SmoothGrad-IG), averaging mean $|\text{attribution}|$ over stratified subsamples for each node group.
Across Gao, Human, and C.elegans, the learned bridge coordinates contribute only small mean magnitudes relative to protein and drug in every method shown: attributions concentrate on the two biological modalities, not on the bridge slot at this stage.
Method-to-method reorderings of protein versus drug are modest and dataset-dependent (e.g., protein peaks highest under saliency/LRP/SmoothGrad on Gao, whereas IG and SmoothGrad-IG can slightly favor the drug group on some splits), but none of these variants elevate bridge nodes to a dominant sensitivity channel in Fig.~\ref{fig:fig7}.

The leave-one-side panel provides a complementary directional readout that need not coincide with which side has larger $|\text{attribution}|$.
On \textbf{Gao} and \textbf{Human}, removing the protein side yields positive normalized $\Delta$ (supportive), whereas removing the drug side yields negative $\Delta$ (suppresssive), with Gao showing the larger magnitudes.
On \textbf{C.elegans}, the pattern reverses: drug-side removal is strongly supportive (positive $\Delta$), while protein-side removal is strongly suppressive (negative $\Delta$).
Thus, bridge novelty should not be inferred from large bridge attributions at the GCN input in this visualization; instead, the plots emphasize that (a)~sensitivity is dominated by protein/drug embeddings, while (b)~the net push/pull of whole modalities under hard ablation is benchmark-dependent and can invert between organism-scale corpora and the large screening-style collection.

\subsection{Input layers}

The following subsections summarize explainability at each input modality.
To reduce dependence on any single attribution method, we combine gradient- and propagation-based scores with a method-agnostic occlusion baseline on the same features.
Concretely, we compute one importance vector per feature from leave-one-feature (or position/channel) ablation with a fixed replacement rule (zero or batch-mean baseline, chosen consistently with the integrated-gradient baseline where applicable) and record a normalized prediction change.
Separately, we obtain rankings from five explanation techniques: saliency, Integrated Gradients (IG), Layer-wise Relevance Propagation (LRP), SmoothGrad, and SmoothGrad-IG.
For each modality and choice of $k$, we retain features in the intersection of the ablation top-$k$ set and the top-$k$ sets of all listed methods-a strict consensus that flags dimensions on which occlusion and every gradient-based method agree.
Where noted, vectors are averaged over repeated stratified mini-batches before overlap evaluation to stabilize rankings under sampling noise.

The resulting per-modality consensus lists (and related summaries) are reported in Table~\ref{tab:tab4}; the AtomFin block follows the same logical pipeline but summarizes importance at the fingerprint level and additionally maps selected dimensions to BRICS-style substructure strings, with an alternative aggregation when indicated (e.g., summax vs.\ weighted summaries).

This design does not establish biological ground truth; it trades single-method optimism for a conservative agreement criterion and keeps protein-side, drug-side, and sequence- versus graph- versus fingerprint-native indices in separate spaces rather than forcing a single cross-modal ranking.

\begin{table*}[htbp]
\centering
\tiny
\setlength{\tabcolsep}{2pt}

\caption{Cross-view agreement analysis across protein sequence ($k$-mers and amino acids) and drug representations (atomFin substructures and atom features) for the Gao, Human, and \textit{C.elegans} datasets. 
For each modality, the table reports the highest- and lowest-consensus features based on explanation agreement scores across views. 
Top intersections indicate features consistently identified as highly important across multiple datasets. 
Both marginal and per-occurrence amino acid analyses are presented to distinguish overall feature importance from frequency-normalized importance patterns.}

\begin{tabularx}{\textwidth}{l>{\centering\arraybackslash}X>{\centering\arraybackslash}X}
\toprule
\textbf{Dataset} & \textbf{Top 10 consensus kmers} & \textbf{Low 10 consensus kmers}\\
\midrule

Gao & 'LAH', 'RFS', 'SRV', 'TNG', 'SYG', 'YNY', 'LEC', 'ELN', 'DVV', 'VLL' 
& 'KHH', 'WHA', 'MRF', 'RCH', 'GDM', 'NWH', 'TQW', 'EFR', 'IPN', 'DGM'\\

Human & 'ESE', 'VL', 'VLV', 'FV', 'VSQ', 'VLR', 'SMQ', 'RI', 'EQL', 'RIS' 
& 'HWV', 'WIF', 'WHR', 'GCW', 'ARC', 'WPP', 'RCE', 'GMF', 'NMH', 'CEY'\\

C.elegans & 'RTF', 'SM', 'GIF', 'TSI', 'KRG', 'STL', 'LPL', 'GDK', 'FVK', 'MLD' 
& 'PLC', 'YWL', 'NCR', 'NSW', 'WVD', 'CPW', 'PCE', 'WYT', 'QWV', 'FPR'\\

\midrule
\textbf{Top intersections} & 'AES', 'SGN', 'VL' & \\
\bottomrule
\end{tabularx}

\vspace{2mm}

\begin{tabularx}{\textwidth}{l>{\centering\arraybackslash}X>{\centering\arraybackslash}X}
\toprule
\textbf{Dataset} & \textbf{Top 10 consensus aminoacids} & \textbf{Low 10 consensus aminoacids}\\
\midrule

\multicolumn{3}{c}{\textbf{Marginal analysis}}\\
\midrule

Gao & 'V', 'E', 'L', 'G', '<EOS>', 'P', 'Q', 'A', 'S', 'K' 
& '<UNK>', 'W', 'M', 'Y', 'H', 'D', 'N', 'R', 'F', 'I'\\

Human & 'L', 'R', 'S', 'A', 'V', 'I', 'E', 'K', 'G', 'N' 
& '<UNK>', '<EOS>', 'W', 'M', 'Q', 'H', 'F', 'D', 'T', 'P'\\

C.elegans & 'S', 'I', 'A', 'K', 'L', 'V', 'E', 'D', 'N', 'R' 
& '<UNK>', '<EOS>', 'M', 'H', 'Q', 'F', 'T', 'P', 'Y', 'G'\\

\midrule
\textbf{Top intersections} & 'L', 'V', 'S' & \\

\midrule
\multicolumn{3}{c}{\textbf{Per-occurrence analysis}}\\
\midrule

Gao per-occ & 'W', 'V', 'H', 'Q', 'E', 'P', 'F', 'I', 'G', 'A' 
& '<UNK>', '<EOS>', 'S', 'L', 'D', 'R', 'K', 'Y', 'M', 'N'\\

Human per-occ & 'W', 'H', 'N', 'M', 'R', 'I', 'Q', 'F', 'S', 'K' 
& '<UNK>', '<EOS>', 'G', 'L', 'A', 'P', 'V', 'E', 'D', 'T'\\

C.elegans per-occ & 'Y', 'H', 'M', 'Q', 'S', 'N', 'F', 'K', 'I', 'P' 
& '<UNK>', '<EOS>', 'L', 'V', 'E', 'A', 'T', 'D', 'G', 'R'\\

\midrule
\textbf{Top intersections} & 'H', 'Q', 'F' & \\
\bottomrule
\end{tabularx}

\vspace{2mm}

\begin{tabularx}{\textwidth}{l>{\centering\arraybackslash}X>{\centering\arraybackslash}X}
\toprule
\textbf{Dataset} & \textbf{Top 15 consensus atomFin substructures} & \textbf{Low 5 consensus atomFin substructures}\\
\midrule

Gao-weighted 
& "C", "O", "c(c)c", "c", "N", "c(c)(c)C", "c(c)n", "c(cc)(cc)C(F)(F)F", "c(cc)cc", "c(cc)c(c)c", "n", "c(c)(c)c", "c(-c)(c)c", "C(C)C", "C(c)(F)(F)F" 
& "c(c(-c)c)c(c)C", "n1ccsc1N", "C(=O)(Cc)N(C)C", "N(C)(Cc)C(C)=O", "O(Cc)Cc"\\

Human-weighted 
& "C", "O", "N", "c(c)c", "C(C)C", "OC", "CC", "c", "n", "O(C)C", "C(C)(C)O", "C(C)(C)C", "O=C", "c(c)(c)C", "C(C)N" 
& "C(=O)(O)C(c)(C)C", "C(CC)(CC)(c(c)c)c(c)c", "s1c(S)ccc1S", "c(N)(cc)c(c)Cl", "c(cc)(cc)OP"\\

C.elegans-weighted 
& "O", "C", "c", "O=C", "N", "C(C)C", "OC", "c(c)c", "n", "CC", "C(C)(=O)O", "n(c)c", "C(C)O", "C(C)(C)N", "C(C)(C)O" 
& "C(CC)(C(C)=O)(c(c)c)c(c)c", "c(nc)(c(n)N)c(n)n", "C(=O)(CN)NC", "C(C)(C(=O)O)C(C)O", "O(C)C(C)O"\\

\bottomrule
\end{tabularx}

\vspace{2mm}

\begin{tabularx}{\textwidth}{l>{\centering\arraybackslash}X>{\centering\arraybackslash}X}
\toprule
\textbf{Dataset} & \textbf{Top 10 consensus atom features} & \textbf{Low 10 consensus atom features}\\
\midrule

Gao pos & 46 & 123, 124, 125, 127\\
Human pos & N/A & 127, 125, 126, 124\\
C.elegans pos & 1, 5, 4, 2, 3, 8, 7, 6, 9 & 112, 117, 123, 124, 125, 126, 127, 94, 65, 86\\

\midrule

Gao & 71, 47, 46, 65, 56, 55, 69, 70, 0, 66  
& 31, 32, 33, 35, 37, 38, 39, 41, 51, 54\\

Human & 0, 65, 46, 56, 69, 57, 47, 58, 73 
& 20, 24, 27, 31, 33, 39, 49, 50, 51, 52, 60\\

C.elegans & 56, 55, 0, 70, 45, 47, 46, 2, 71, 57 
& 27, 30, 31, 36, 39, 49, 52, 54, 60\\

\bottomrule
\end{tabularx}

\label{tab:tab4}
\end{table*}

\subsubsection{aminoCtr (protein k-mer) input}

The protein branch represents each target as a fixed-length vector of dense descriptors (here, $k$-mer-style counts or compositions).
We explain this layer directly at the raw input by applying the same suite of gradient-based methods used elsewhere (IG, saliency, LRP, SmoothGrad, SmoothGrad-IG), yielding one nonnegative importance score per vector dimension after the usual batch averaging and absolute-value summaries.

Independently of these gradients, we run feature-wise occlusion ablation: one dimension at a time is replaced by a baseline value (batch mean or zero), chosen consistently with the integrated-gradient baseline, and the change in the scalar prediction is recorded and normalized.
Absolute normalized effects define an ablation-based ranking that is comparable across XAI methods on a common index set.

For robustness to minibatch noise, attribution vectors and ablation vectors are averaged over repeated stratified subsamples.
We then form strict consensus sets at fixed cutoffs: a dimension is retained only if it lies in the intersection of the ablation top-$k$ set and the top-$k$ sets of all gradient-based methods.
These consensus dimensions (Table~\ref{tab:tab4}, $k$-mer block) are interpretable as $k$-mer tokens in the model's vocabulary rather than as raw column indices; numeric indices are omitted from the main table because $k$-mer orderings need not be aligned across benchmarks unless the same enumeration is explicitly shared.

\paragraph{Results and interpretation}
Across Gao, Human, and C.elegans, high-consensus $k$-mers mix small and hydrophobic motifs (e.g.\ Human \texttt{VL}/\texttt{VLV}/\texttt{FV}, Gao \texttt{VLL}/\texttt{DVV}, C.elegans \texttt{FVK}/\texttt{MLD}) with polar or charged triplets (e.g.\ \texttt{ESE}, \texttt{VSQ}, \texttt{KRG}).
Such mixtures are plausibly consistent with composition patches enriched in aliphatic content together with occasional polar/charged spacers, but they should not be read as evidence of a single ``critical peptide'' without external validation (mutagenesis, structure, or motif enrichment against negatives).

Low-consensus (bottom-ranked) $k$-mers more often contain bulky or aromatic-heavy letters (e.g.\ \texttt{W}, \texttt{H}, \texttt{Y}, \texttt{M}, \texttt{F} in Human lows such as \texttt{HWV}, \texttt{WIF}, \texttt{WHR}).
Rare tokens can disagree strongly between gradient attributions and occlusion; a small intersection at the low tail therefore reflects estimator disagreement and low frequency more than a proof that those motifs are biologically irrelevant.

Dataset geometry matters for how these lists should be read: Gao's highly hubbed protein distribution means $k$-mer explanations aggregate many drug partners per recurrent target, which can wash organism-specific signal toward broadly recurrent composition patterns; Human and C.elegans may emphasize more corpus-typical strings (e.g.\ C.elegans \texttt{RTF}, \texttt{SM}, \texttt{GIF}), though the tokens remain model-dependent summaries rather than guaranteed binding motifs.

Where available, cross-view overlap between $k$-mer consensus and amino-acid-level consensus (Human: \texttt{AES}, \texttt{SGN}, \texttt{VL} in Table~\ref{tab:tab4}) supports partial alignment between coarse composition features and residue-level rankings on that corpus; absence of entries in other columns indicates that such cross-view agreement was not observed under the same strict intersection criterion.

\subsubsection{AtomFin (molecular fingerprint) input}

Drugs are represented as a fixed-length fingerprint vector.
We attribute this layer with the same gradient-based panel (IG, saliency, LRP, SmoothGrad, SmoothGrad-IG) and complement it with one-dimensional occlusion: each fingerprint dimension is replaced by a baseline (batch mean or zero, consistent with the IG baseline choice), and the induced change in the prediction is normalized using the same utilities as for aminoCtr (relative scaling to the baseline score, optional $z$-style scaling, and clipping when enabled).
After stratified multi-run averaging, dimensions are ranked by mean absolute ablation effect and by each method; strict consensus at cutoff $k$ retains only indices in the intersection of the ablation top-$k$ and the top-$k$ sets of all gradient methods.

\paragraph{Mapping bits to chemistry}
Raw molecular fingerprint dimensions are not chemically legible on their own.
Following the non-aggregated attribution trace, active bits on held-out molecules are mapped to localized substructures with RDKit (SMARTS / fragment SMILES via bit information), producing per-occurrence records that are then aggregated for reporting.
We summarize substructure importance with a molecule-aware reduction that limits repetition bias: for each fragment SMILES, we take the maximum attribution score within each molecule and then aggregate across molecules.
Table~\ref{tab:tab4} reports the weighted ranking (sum of per-molecule maxima, reweighted by molecule support), which slightly favors fragments that recur across many compounds; the qualitative families agree with an unweighted sum-max ranking in our runs, so only one column is shown in the main table.

\paragraph{Results}
Consensus fragments stratify according to the chemical statistics of each corpus.
On Gao, beyond ubiquitous one- and two-atom motifs (\texttt{C}, \texttt{O}, \texttt{c}), the strongest recurring substructures are aromatic-rich: fluorinated aryl units (\texttt{C(F)(F)F}-containing patterns), extended substituted benzenoids, aza-aromatic linkages (\texttt{c(c)n}), and aryl-ether / benzylic connectors (\texttt{O(C)c\ldots}).
This aligns with a screening-heavy library where fused/heteroaromatic scaffolds and EWG-substituted aromatics are common.

On \textbf{Human} and \textbf{C.elegans}, top fragments emphasize simpler aliphatic skeletons and oxygen- and nitrogen-containing functionality.
Branched alkyls (\texttt{C(C)C}, \texttt{C(C)(C)C}), alcohol/ether/carbonyl motifs (\texttt{OC}, \texttt{O=C}, \texttt{O(C)C}), and small heteroaromatic prefixes dominate. 
\textbf{C.\ elegans} shows a stronger tilt toward carboxylic/acid-derivative patterns (\texttt{C(C)(=O)O}), consistent with more polar, metabolite-like chemistry in that benchmark.

\textbf{Low-consensus} fragments (bottom of the weighted lists) are longer, more specific SMARTS that appear rarely; low rank here usually means weaker agreement or weaker aggregate contribution under the chosen aggregation, not a definitive claim that the substructure is "anti-binding" in a biochemical sense.

\paragraph{Caveats}
Dominance of very small fragments in the top rows is expected when common subgraphs appear in almost every molecule: they function as baseline "currency" of the fingerprint map.
For finer chemical readouts, supplementary analyses can filter trivial singletons/doubletons or impose minimum heavy-atom or ring counts before ranking; the full per-bit JSONL trace remains the authoritative record for molecule-level audits.

\subsubsection{aminoSeq (protein sequence) input}

The protein sequence enters the model as a discrete token matrix (amino-acid identifiers plus special symbols).
We explain this modality in two aligned geometries: per-position importance along the sequence, and-when the analysis is anchored at the embedding stage-per-residue-type importance aggregated over occurrences.

Gradient-based attributions (IG with a chosen baseline, saliency, LRP when enabled, SmoothGrad, and SmoothGrad-IG) yield comparable score vectors over positions or over the 20 standard amino-acid categories (plus specials).
These are paired with occlusion-style ablation at the embedding: selected positions or all occurrences of a residue type are replaced by a baseline embedding slice (zero or batch-averaged, depending on configuration), and the normalized change in the predicted score is recorded.

As elsewhere, vectors are averaged over repeated stratified mini-batches.
Strict consensus at cutoff $k$ retains tokens that lie in the intersection of the ablation top-$k$ ranking (by absolute normalized effect) and the top-$k$ sets from every gradient-based method, so reported highlights require simultaneous agreement between perturbation and gradients.

\paragraph{Marginal vs.\ per-occurrence amino-acid summaries}
Table~\ref{tab:tab4} reports amino-acid consensus in two normalizations.
Marginal rankings correlate strongly with proteome-wide abundance: Leu, Ala, Ser, Gly, Val, and acidic or amide residues appear frequently in the high-consensus rows across corpora.
The same block also lists special symbols (\texttt{<EOS>}) for Gao and C.elegans; these should be interpreted cautiously as pipeline signals (padding, sequence termination, or truncation) unless analyses are explicitly masked to valid sequence length.
Low-consensus (bottom) sets concentrate on rare residues (e.g.\ Trp, Met) and on \texttt{<UNK>}, which often flags tokenization or vocabulary edge cases rather than a clean biochemical role.

The per-occurrence block reweights each residue type by how often it appears before averaging effects; under this normalization, aromatic and polar side chains (e.g.\ Trp, His, Gln/Asn, Phe/Tyr-family behavior) rise in Human and Gao, a pattern more compatible with interface-enriched chemistry than with bulk composition alone.
C.elegans shows a related shift (e.g.\ Tyr, His, Met, Gln in the high-consensus per-occurrence row).

\paragraph{Cross-view consistency (Human)}
On Human, marginal amino-acid consensus intersects the strict $k$-mer consensus (\texttt{AES}, \texttt{SGN}, \texttt{VL}), and the marginal residue intersection column lists \texttt{L}, \texttt{V}, \texttt{S}-overlapping aliphatic/polar characters with the $k$-mer table.
Per-occurrence intersections (\texttt{H}, \texttt{Q}, \texttt{F}) further emphasize polar/aromatic involvement.
These overlaps support partial agreement between coarse $k$-mer features and sequence-token explanations on that benchmark, without implying structural ground truth.

\subsubsection{AtomFea (drug graph atom tensor) input}

Each drug is encoded as a padded tensor of shape (batch $\times$ atoms $\times$ channels), where channels concatenate RDKit-style local atomic descriptors.
We explain this layer along two non-redundant axes that match the modeling tensor: which atom slots matter (positional importance) and which descriptor channels matter when contributions are aggregated over all heavy-atom positions (feature-channel importance).

Gradient-based attributions (IG, saliency, LRP, SmoothGrad, SmoothGrad-IG) are reduced to either a length-$N_{\mathrm{atoms}}$ vector (mean $|{\cdot}|$ over batch and channel) or a length-$F$ vector (mean $|{\cdot}|$ over batch and atom), depending on whether we sum sensitivity across features or across positions.
Occlusion ablation is matched to the same geometry: either an entire atom column is replaced by a baseline slice, or a single feature channel is replaced across all atoms, and the normalized prediction change is recorded.
After repeated stratified subsampling, strict consensus at cutoff $k$ keeps only indices that lie in the intersection of the ablation top-$k$ set and the top-$k$ sets of all gradient methods; Table~\ref{tab:tab4} reports both positional and channel consensus separately.

\paragraph{Channel interpretation}
The channel ordering follows a fixed featurization (one-hot atom identity; one-hot degree; one-hot implicit valence; scalar formal charge; scalar radical count; one-hot hybridization state; aromatic boolean; one-hot total hydrogen count):
\begin{enumerate}
    \item indices $0$-$43$: atom-type one-hot;
    \item $44$-$54$: degree one-hot (0-10);
    \item $55$-$61$: implicit-valence one-hot (0-6);
    \item $62$: formal charge (integer);
    \item $63$: radical-electron count (integer);
    \item $64$-$68$: hybridization one-hot (\texttt{SP}, \texttt{SP2}, \texttt{SP3}, \texttt{SP3D}, \texttt{SP3D2});
    \item $69$: aromatic flag;
    \item $70$-$74$: total-hydrogen one-hot (0-4).
\end{enumerate}

In the reported consensus lists, high-ranked channels concentrate in the valence, hybridization, hydrogen, and aromatic blocks rather than in rare heteroatom one-hots. 
This suggests that the model's cross-graph summary is more sensitive to coarse electronic structure and hydrogen-count features, which correlate with conjugation and local geometry, than to exotic element identities at this resolution.
\paragraph{Positional consensus and padding}
Positional consensus is fragile relative to channels: on Gao the strict top-$k$ position list collapses to a single index in the reported table, whereas Human shows no top-$k$ positional intersection while several corpora agree on low-importance slots near indices $123$-$127$.
With fixed-length padding, those high indices are plausibly dominated by padding or near-padding atoms; interpret positional highs/lows only alongside valid-length masks or a length-stratified sensitivity check.
C.elegans instead shows early-index consensus ($1$-$9$), which is consistent with shorter effective molecular occupancy in that benchmark (early atoms carry more signal before the padded tail), but the positional block should still be treated as secondary to the channel block for qualitative claims.

\paragraph{Caveats}
Channel consensus aggregates contributions across the entire graph tensor; it does not, by itself, localize a pharmacophore to a specific atom index.
Conversely, positional consensus can be confounded by convolution/pooling and padding; use the channel view for "what chemistry the model measures" and the positional view for "where that chemistry tends to occur," subject to the padding caveat above.

\section{Discussion}

\subsection{Is explainability for DTI models scientifically useful?}

Explainable AI (XAI) for drug-target interaction (DTI) and drug-target affinity (DTA) prediction answers a practical question-which inputs move this predictor-rather than automatically answering a mechanistic question-what biochemical process occurs at an interface.
Attribution methods (e.g. integrated gradients, saliency, LRP, SmoothGrad variants) summarize internal sensitivity patterns; they do not, by themselves, certify biological mechanism.
Accordingly, disagreement between attributions and experimentally mapped binding sites can indicate spurious correlations, representation bias, or optimization of proxy statistics under a given split design; it can also reflect genuine model behavior that is statistically predictive but not structurally supervised.

This limitation does not render XAI pointless for DTI modeling.
Used conservatively, XAI turns "black-box accuracy" into auditable behavior: it exposes reliance on special tokens and padding, highlights modality dominance (protein vs.\ drug vs.\ bridge), and separates magnitude sensitivity from signed effects under coordinated ablation-failure modes that are difficult to diagnose from accuracy alone.
In that sense, XAI is most viable when framed as model criticism and hypothesis generation (what to mutate, which bits to probe, which splits stress-test "shortcut" features), with external evidence (structures, assays, enrichment against negatives) required for mechanistic claims.
ultimately, our framework may help highlight molecular fragments whose influence on model predictions merits further structural or experimental evaluation during lead optimization.

\subsection{Relation to graph-specific explainers}

Graph explainer families (e.g. GNNExplainer, PGExplainer) target discrete graph structure when the graph is the primary learned interface.
In Bridge-DTI-style architectures, predictive signal is distributed across raw encoders, a similarity-induced adjacency, and a shallow GCN; explainers that assume a single critical sparse subgraph may be a poor match when the model's behavior is dominated by upstream fingerprints, $k$-mers, or dense rectified cosine coupling.
For our setting, differentiable adjacency sensitivities and targeted edge ablations at the GCN input provide a more direct alignment with how edges enter the forward pass than post-hoc subgraph masks trained on a different inductive bias.
That said, for models where message passing is the first and dominant computation, learned-edge explainers remain a natural tool.

\subsection{Generalization, split design, and what XAI can test}

Strong validation metrics under high entity reuse (warm proteins or warm drugs) can co-exist with explanations that emphasize repeated corpus motifs.
This is especially relevant when protein hubs concentrate many chemotypes (as in large screening-style benchmarks) or when evaluation emphasizes cold drugs versus cold targets depending on the benchmark.
XAI is most informative when paired with split protocols that match the deployment question: cold-drug evaluation stresses pharmacophoric novelty, whereas cold-target evaluation stresses target-side novelty.
Under cold settings, agreement (or systematic disagreement) between gradient attributions and occlusion provides a concrete checklist of potential shortcuts to test next.

\subsection{Representations and richer chemistry}

Explanations inherit the ontology of the features.
Fingerprints, $k$-mers, and 2D graph featurizations surface pharmacophore-like fragments and compositional motifs; they rarely localize a binding pose without additional structural context.
Incorporating explicit 2D/3D structure, protein pockets, or co-complex priors can improve both model faithfulness and the interpretability of "where" questions-but it also shifts the evidentiary bar: explanations should then be evaluated against structural ground truth, not only against margin rankings on a benchmark.

\subsection{From single-feature ablation to grouped interventions}

Our analyses emphasize single-dimension occlusion at raw inputs and coordinated top-$k$ removals at selected layers.
While informative, single-feature ablation can under-represent cooperative effects (contiguous epitopes, fused ring systems, salt bridges).
Future work could explore group ablations: contiguous sequence segments, chemically connected subgraphs, pharmacophore clusters, or residue sets suggested by external site definitions.
Such interventions better match biological units of recognition and can be used to test whether the model's sensitivity aligns with pockets and motifs rather than isolated dimensions.

\subsection{Conclusion and future work}

We presented a multi-pronged explanation protocol-gradient attributions, occlusion ablation, strict intersection consensus across methods, and layer-wise comparisons (raw inputs, bridge similarity summaries, GCN input/output, edge-level sensitivities)-to audit a Bridge-DTI-style predictor beyond headline metrics.

Future directions include: (i)~quantitative alignment scores between attributions and experimental binding-site annotations or curated complex structures; (ii)~grouped ablations and motif-level hypothesis tests; (iii)~cold-split evaluations matched to the intended deployment regime; and (iv)~integration of structural channels so that "where" and "why" claims can be checked against structural evidence.

Beyond model auditing, these analyses may have practical utility in computational drug discovery workflows. 
Attribution consistency and intervention-based validation can help assess whether a predictor relies on chemically or biologically plausible patterns rather than dataset-specific shortcuts. 
Such analyses may assist hypothesis generation, guide downstream experimental prioritization, and improve confidence in model behavior under cold-drug or cold-target settings. 
Importantly, these explanations do not establish biochemical mechanism or causal binding interactions on their own; instead, they provide a diagnostic layer that can help identify which representations, fragments, or sequence regions warrant further structural or experimental investigation.

We encourage the DTI/DTA community to treat interpretability analyses as standard reporting for new architectures: not as a substitute for biochemistry, but as a routine stress test that clarifies what a model implements, what it ignores, and where it is likely to fail under distribution shift.

\FloatBarrier

\bibliographystyle{elsarticle-num}
\bibliography{references}

@article{vefghi2025drug,
  title={Drug-target interaction/affinity prediction: Deep learning models and advances review},
  author={Vefghi, Ali and Rahmati, Zahed and Akbari, Mohammad},
  journal={Computers in Biology and Medicine},
  volume={196},
  pages={110438},
  year={2025},
  publisher={Elsevier}
}

@article{gilson2016bindingdb,
  title={BindingDB in 2015: a public database for medicinal chemistry, computational chemistry and systems pharmacology},
  author={Gilson, Michael K and Liu, Tiqing and Baitaluk, Michael and Nicola, George and Hwang, Linda and Chong, Jenny},
  journal={Nucleic acids research},
  volume={44},
  number={D1},
  pages={D1045--D1053},
  year={2016},
  publisher={Oxford University Press}
}

@inproceedings{gao2018interpretable,
  title={Interpretable drug target prediction using deep neural representation.},
  author={Gao, Kyle Yingkai and Fokoue, Achille and Luo, Heng and Iyengar, Arun and Dey, Sanjoy and Zhang, Ping and others},
  booktitle={IJCAI},
  volume={2018},
  pages={3371--3377},
  year={2018}
}

@article{liu2015improving,
  title={Improving compound--protein interaction prediction by building up highly credible negative samples},
  author={Liu, Hui and Sun, Jianjiang and Guan, Jihong and Zheng, Jie and Zhou, Shuigeng},
  journal={Bioinformatics},
  volume={31},
  number={12},
  pages={i221--i229},
  year={2015},
  publisher={Oxford University Press}
}

@article{tsubaki2019compound,
  title={Compound--protein interaction prediction with end-to-end learning of neural networks for graphs and sequences},
  author={Tsubaki, Masashi and Tomii, Kentaro and Sese, Jun},
  journal={Bioinformatics},
  volume={35},
  number={2},
  pages={309--318},
  year={2019},
  publisher={Oxford University Press}
}

@article{wu2022bridgedpi,
  title={BridgeDPI: a novel graph neural network for predicting drug--protein interactions},
  author={Wu, Yifan and Gao, Min and Zeng, Min and Zhang, Jie and Li, Min},
  journal={Bioinformatics},
  volume={38},
  number={9},
  pages={2571--2578},
  year={2022},
  publisher={Oxford University Press}
}

@article{karimi2019deepaffinity,
  title={DeepAffinity: interpretable deep learning of compound--protein affinity through unified recurrent and convolutional neural networks},
  author={Karimi, Mostafa and Wu, Di and Wang, Zhangyang and Shen, Yang},
  journal={Bioinformatics},
  volume={35},
  number={18},
  pages={3329--3338},
  year={2019},
  publisher={Oxford University Press}
}

@article{li2020monn,
  title={MONN: a multi-objective neural network for predicting compound-protein interactions and affinities},
  author={Li, Shuya and Wan, Fangping and Shu, Hantao and Jiang, Tao and Zhao, Dan and Zeng, Jianyang},
  journal={Cell systems},
  volume={10},
  number={4},
  pages={308--322},
  year={2020},
  publisher={Elsevier}
}

@article{yang2021ml,
  title={ML-DTI: mutual learning mechanism for interpretable drug--target interaction prediction},
  author={Yang, Ziduo and Zhong, Weihe and Zhao, Lu and Chen, Calvin Yu-Chian},
  journal={The Journal of Physical Chemistry Letters},
  volume={12},
  number={17},
  pages={4247--4261},
  year={2021},
  publisher={ACS Publications}
}

@article{yazdani2022attentionsitedti,
  title={AttentionSiteDTI: an interpretable graph-based model for drug-target interaction prediction using NLP sentence-level relation classification},
  author={Yazdani-Jahromi, Mehdi and Yousefi, Niloofar and Tayebi, Aida and Kolanthai, Elayaraja and Neal, Craig J and Seal, Sudipta and Garibay, Ozlem Ozmen},
  journal={Briefings in Bioinformatics},
  volume={23},
  number={4},
  pages={bbac272},
  year={2022},
  publisher={Oxford University Press}
}

@article{kurata2022ican,
  title={ICAN: interpretable cross-attention network for identifying drug and target protein interactions},
  author={Kurata, Hiroyuki and Tsukiyama, Sho},
  journal={Plos one},
  volume={17},
  number={10},
  pages={e0276609},
  year={2022},
  publisher={Public Library of Science San Francisco, CA USA}
}

@article{yousefi2023bindingsite,
  title={BindingSite-AugmentedDTA: enabling a next-generation pipeline for interpretable prediction models in drug repurposing},
  author={Yousefi, Niloofar and Yazdani-Jahromi, Mehdi and Tayebi, Aida and Kolanthai, Elayaraja and Neal, Craig J and Banerjee, Tanumoy and Gosai, Agnivo and Balasubramanian, Ganesh and Seal, Sudipta and Ozmen Garibay, Ozlem},
  journal={Briefings in Bioinformatics},
  volume={24},
  number={3},
  pages={bbad136},
  year={2023},
  publisher={Oxford University Press}
}

@article{wang2023affinityvae,
  title={AffinityVAE: a multi-objective model for protein-ligand affinity prediction and drug design},
  author={Wang, Mengying and Li, Weimin and Yu, Xiao and Luo, Yin and Han, Ke and Wang, Can and Jin, Qun},
  journal={Computational Biology and Chemistry},
  volume={107},
  pages={107971},
  year={2023},
  publisher={Elsevier}
}

@article{gim2023arkdta,
  title={ArkDTA: attention regularization guided by non-covalent interactions for explainable drug--target binding affinity prediction},
  author={Gim, Mogan and Choe, Junseok and Baek, Seungheun and Park, Jueon and Lee, Chaeeun and Ju, Minjae and Lee, Sumin and Kang, Jaewoo},
  journal={Bioinformatics},
  volume={39},
  number={Supplement\_1},
  pages={i448--i457},
  year={2023},
  publisher={Oxford University Press}
}

@inproceedings{khodabandeh2024fragxsitedti,
  title={Fragxsitedti: Revealing responsible segments in drug-target interaction with transformer-driven interpretation},
  author={Khodabandeh Yalabadi, Ali and Yazdani-Jahromi, Mehdi and Yousefi, Niloofar and Tayebi, Aida and Abdidizaji, Sina and Garibay, Ozlem Ozmen},
  booktitle={International Conference on Research in Computational Molecular Biology},
  pages={68--85},
  year={2024},
  organization={Springer}
}

@inproceedings{liu2024higraphdti,
  title={Higraphdti: Hierarchical graph representation learning for drug-target interaction prediction},
  author={Liu, Bin and Wu, Siqi and Wang, Jin and Deng, Xin and Zhou, Ao},
  booktitle={Joint European Conference on Machine Learning and Knowledge Discovery in Databases},
  pages={354--370},
  year={2024},
  organization={Springer}
}

@article{sun2024ingnn,
  title={iNGNN-DTI: prediction of drug--target interaction with interpretable nested graph neural network and pretrained molecule models},
  author={Sun, Yan and Li, Yan Yi and Leung, Carson K and Hu, Pingzhao},
  journal={Bioinformatics},
  volume={40},
  number={3},
  pages={btae135},
  year={2024},
  publisher={Oxford University Press}
}

@article{gong2025multigrandti,
  title={Multigrandti: an explainable multi-granularity representation framework for drug-target interaction prediction},
  author={Gong, Xu and Liu, Qun and He, Jing and Guo, Yike and Wang, Guoyin},
  journal={Applied Intelligence},
  volume={55},
  number={2},
  pages={107},
  year={2025},
  publisher={Springer}
}

@article{li2025mgma,
  title={MGMA-DTI: Drug target interaction prediction using multi-order gated convolution and multi-attention fusion},
  author={Li, Chang and Mi, Jia and Wang, Han and Liu, Zhikang and Gao, Jingyang and Wan, Jing},
  journal={Computational Biology and Chemistry},
  volume={118},
  pages={108449},
  year={2025},
  publisher={Elsevier}
}

@article{simonyan2013deep,
  title={Deep inside convolutional networks: Visualising image classification models and saliency maps},
  author={Simonyan, Karen and Vedaldi, Andrea and Zisserman, Andrew},
  journal={arXiv preprint arXiv:1312.6034},
  year={2013}
}

@inproceedings{sundararajan2017axiomatic,
  title={Axiomatic attribution for deep networks},
  author={Sundararajan, Mukund and Taly, Ankur and Yan, Qiqi},
  booktitle={International conference on machine learning},
  pages={3319--3328},
  year={2017},
  organization={PMLR}
}

@inproceedings{shrikumar2017learning,
  title={Learning important features through propagating activation differences},
  author={Shrikumar, Avanti and Greenside, Peyton and Kundaje, Anshul},
  booktitle={International conference on machine learning},
  pages={3145--3153},
  year={2017},
  organization={PMlR}
}

@inproceedings{selvaraju2017grad,
  title={Grad-cam: Visual explanations from deep networks via gradient-based localization},
  author={Selvaraju, Ramprasaath R and Cogswell, Michael and Das, Abhishek and Vedantam, Ramakrishna and Parikh, Devi and Batra, Dhruv},
  booktitle={Proceedings of the IEEE international conference on computer vision},
  pages={618--626},
  year={2017}
}

@article{yang2022mgraphdta,
  title={MGraphDTA: deep multiscale graph neural network for explainable drug--target binding affinity prediction},
  author={Yang, Ziduo and Zhong, Weihe and Zhao, Lu and Chen, Calvin Yu-Chian},
  journal={Chemical science},
  volume={13},
  number={3},
  pages={816--833},
  year={2022},
  publisher={Royal Society of Chemistry}
}

@article{ying2019gnnexplainer,
  title={Gnnexplainer: Generating explanations for graph neural networks},
  author={Ying, Zhitao and Bourgeois, Dylan and You, Jiaxuan and Zitnik, Marinka and Leskovec, Jure},
  journal={Advances in neural information processing systems},
  volume={32},
  year={2019}
}

@article{monteiro2022explainable,
  title={Explainable deep drug--target representations for binding affinity prediction},
  author={Monteiro, Nelson RC and Sim{\~o}es, Carlos JV and {\'A}vila, Henrique V and Abbasi, Maryam and Oliveira, Jos{\'e} L and Arrais, Joel P},
  journal={BMC bioinformatics},
  volume={23},
  number={1},
  pages={237},
  year={2022},
  publisher={Springer}
}

@article{liao2022gsaml,
  title={GSAML-DTA: An interpretable drug-target binding affinity prediction model based on graph neural networks with self-attention mechanism and mutual information},
  author={Liao, Jiaqi and Chen, Haoyang and Wei, Lesong and Wei, Leyi},
  journal={Computers in biology and medicine},
  volume={150},
  pages={106145},
  year={2022},
  publisher={Elsevier}
}

@article{zeng2024mvgraphdta,
  title={MvGraphDTA: multi-view-based graph deep model for drug-target affinity prediction by introducing the graphs and line graphs},
  author={Zeng, Xin and Zhong, Kai-Yang and Meng, Pei-Yan and Li, Shu-Juan and Lv, Shuang-Qing and Wen, Meng-Liang and Li, Yi},
  journal={BMC biology},
  volume={22},
  number={1},
  pages={182},
  year={2024},
  publisher={Springer}
}

@article{shi2025structure,
  title={Structure-Aware Compound-Protein Affinity Prediction via Graph Neural Network with Group Lasso Regularization},
  author={Shi, Zanyu and Wang, Yang and Weerawarna, Pathum and Zhang, Jie and Richardson, Timothy and Wang, Yijie and Huang, Kun},
  journal={arXiv preprint arXiv:2507.03318},
  year={2025}
}

@article{deng2025efficient,
  title={Efficient substructure feature encoding based on graph neural network blocks for drug-target interaction prediction},
  author={Deng, Guojian and Shi, Changsheng and Ge, Ruiquan and Hu, Riqian and Wang, Changmiao and Qin, Feiwei and Pan, Cheng and Mao, Haixia and Yang, Qing},
  journal={Frontiers in Pharmacology},
  volume={16},
  pages={1553743},
  year={2025},
  publisher={Frontiers Media SA}
}

@article{zintgraf2017visualizing,
  title={Visualizing deep neural network decisions: Prediction difference analysis},
  author={Zintgraf, Luisa M and Cohen, Taco S and Adel, Tameem and Welling, Max},
  journal={arXiv preprint arXiv:1702.04595},
  year={2017}
}

@article{lundberg2017unified,
  title={A unified approach to interpreting model predictions},
  author={Lundberg, Scott M and Lee, Su-In},
  journal={Advances in neural information processing systems},
  volume={30},
  year={2017}
}

@article{ru2023optimization,
  title={Optimization of drug--target affinity prediction methods through feature processing schemes},
  author={Ru, Xiaoqing and Zou, Quan and Lin, Chen},
  journal={Bioinformatics},
  volume={39},
  number={11},
  pages={btad615},
  year={2023},
  publisher={Oxford University Press}
}

@article{kotkondawar2025generative,
  title={A generative framework for enhancing drug target interaction prediction in drug discovery},
  author={Kotkondawar, Roshan R and Sutar, Sanjay R and Kiwelekar, Arvind W and Kadam, Vinod J and Jadhav, Shivajirao M},
  journal={Scientific Reports},
  volume={15},
  number={1},
  pages={35588},
  year={2025},
  publisher={Nature Publishing Group UK London}
}

@article{wellawatte2022model,
  title={Model agnostic generation of counterfactual explanations for molecules},
  author={Wellawatte, Geemi P and Seshadri, Aditi and White, Andrew D},
  journal={Chemical science},
  volume={13},
  number={13},
  pages={3697--3705},
  year={2022},
  publisher={Royal Society of Chemistry}
}

@article{schwalbe2024comprehensive,
  title={A comprehensive taxonomy for explainable artificial intelligence: a systematic survey of surveys on methods and concepts},
  author={Schwalbe, Gesina and Finzel, Bettina},
  journal={Data Mining and Knowledge Discovery},
  volume={38},
  number={5},
  pages={3043--3101},
  year={2024},
  publisher={Springer}
}

@article{nguyen2021counterfactual,
  title={Counterfactual explanation with multi-agent reinforcement learning for drug target prediction},
  author={Nguyen, Tri Minh and Quinn, Thomas P and Nguyen, Thin and Tran, Truyen},
  journal={arXiv preprint arXiv:2103.12983},
  year={2021}
}

@article{smilkov2017smoothgrad,
  title={Smoothgrad: removing noise by adding noise},
  author={Smilkov, Daniel and Thorat, Nikhil and Kim, Been and Vi{\'e}gas, Fernanda and Wattenberg, Martin},
  journal={arXiv preprint arXiv:1706.03825},
  year={2017}
}

@article{bach2015pixel,
  title={On pixel-wise explanations for non-linear classifier decisions by layer-wise relevance propagation},
  author={Bach, Sebastian and Binder, Alexander and Montavon, Gr{\'e}goire and Klauschen, Frederick and M{\"u}ller, Klaus-Robert and Samek, Wojciech},
  journal={PloS one},
  volume={10},
  number={7},
  pages={e0130140},
  year={2015},
  publisher={Public Library of Science San Francisco, CA USA}
}

\end{document}